\documentclass[letterpaper,twocolumn,10pt]{article}
\usepackage{usenix2019_v3}

\usepackage{tikz}
\usepackage{amsthm} 
\usepackage{amsmath}
\usepackage{amssymb} 
\usepackage{algorithm} 
\usepackage{algorithmic} 
\usepackage{color} 
\usepackage{graphicx} 
\usepackage{subfigure} 
\usepackage{array} 
\usepackage{booktabs} 
\usepackage{multirow} 
\usepackage{bbding} 

\usepackage{color}
\usepackage{tabularx}

\newtheorem{lemma}{Lemma}
\newtheorem{theorem}{Theorem}
\begin{document}

\date{}

\title{\Large \bf Batch Label Inference and Replacement Attacks \\ in Black-Boxed Vertical Federated Learning}

\author{
{\rm Yang Liu\footnotemark[1]}\\
Institute for AI Industry Research, Tsinghua University\\
\and
{\rm Tianyuan Zou\footnotemark[1]}\\
Tsinghua University\\
\and
{\rm Yan Kang}\\
Webank, Shenzhen, China\\
\and
{\rm Wenhan Liu}\\
Shandong University\\
\and
{\rm Yuanqin He}\\
Webank, Shenzhen, China\\
\and
{\rm Zhihao Yi}\\
Webank, Shenzhen, China\\
\and
{\rm Qiang Yang}\\
Webank and Hong Kong University of Science and Technology\\
} 

\maketitle

\footnotetext[1]{The two authors contributed equally to this paper.}

\begin{abstract}
In a vertical federated learning (VFL) scenario where features and model are split into different parties, communications of sample-specific updates are required for correct gradient calculations but can be used to deduce important sample-level label information. An immediate defense strategy is to protect sample-level messages communicated with Homomorphic Encryption (HE), and in this way only the batch-averaged local gradients are exposed to each party (termed \textbf{black-boxed VFL}). In this paper, we first explore the possibility of recovering labels in the vertical federated learning setting with HE-protected communication, and show that private labels can be reconstructed with high accuracy by training a gradient inversion model. Furthermore, we show that label replacement backdoor attacks can be conducted in black-boxed VFL by directly replacing encrypted communicated messages (termed \textbf{gradient-replacement attack}). As it is a common presumption that batch-averaged information is safe to share, batch label inference and replacement attacks are a severe challenge to VFL. To defend against batch label inference attack, we further evaluate several defense strategies, including confusional autoencoder (CoAE), a technique we proposed based on autoencoder and entropy regularization. We demonstrate that label inference and replacement attacks can be successfully blocked by this technique without hurting as much main task accuracy as compared to existing methods. 

\end{abstract}

\section{Introduction}
Federated Learning (FL) \cite{DBLP:journals/corr/McMahanMRA16,Yang-et-al:2019,Kairouz2019AdvancesAO} is a collaborative learning framework for training deep learning models with data privacy protection. In the original proposal of cross-device FL \cite{DBLP:journals/corr/McMahanMRA16}, data samples are distributed among different participants, thus these works can be regarded as sample-partitioned federated learning, or horizontal federated learning (HFL) \cite{Yang-et-al:2019}. However, feature-partitioned federated learning, or vertical federated learning (VFL) \cite{Yang-et-al:2019,Hu2019FDMLAC,Liu2019ACE,liu2018ftl,secureboost,HE2020grouptransfer} is another important scenario for many real-world applications. For example, when a bank and an E-commerce company collaboratively train a credit risk model, different features of the same group of users are partitioned among different parties. It is important to safeguard the VFL framework from attacks and data leakages.

Because FL is a collaboration system that requires parties to exchange gradient or model level information, it has been of great research interest to study the potential information leakage from gradients. In FL scenarios, curious parities might be honest but attempt to infer other parties' private data through inference attacks while malicious parties might manipulate the learning process for their own purposes through backdoor attacks. Previous works \cite{Deepleakage2019,DBLP:journals/corr/abs-2001-02610,Yin_2021_CVPR_batch_gradinversion} have shown that it is possible to recover pixel-level (deep leakage) raw data from transmitted gradients information. Previous studies on label leakage of VFL \cite{label_leakage_VFL_2021} have made the assumption that per-sample gradient is communicated and revealed in VFL, therefore the value of the gradients can be used to infer the label. However, sample-level information is not necessarily available in VFL with Homomorphic Encryption (HE) or Multi-Party Secure Computation (MPC) protection \cite{Yang-et-al:2019}, where the encrypted intermediate results are communicated, and both data and local model parameters are kept secret. To draw an analogy to the white-box and black-box attack in deep learning, we consider this scenario \textbf{black-boxed VFL} whereas the VFL with plain-text communication of sample-level information \textbf{white-boxed VFL}. Therefore attacks and defenses in the black-boxed VFL scenario are more challenging since only batch-leveled local gradients are available. 

\begin{algorithm}
\caption{A vertical federated learning framework with HE encryption}
\textbf{Input}: learning rate $\eta$\\
\textbf{Output}: Model parameters $\theta_1$, $\theta_2$ ... $\theta_K$
\begin{algorithmic}[1]
\STATE Party 1,2,\dots,$K$, initialize $\theta_1$, $\theta_2$, ... $\theta_K$. \\
\STATE TTP creates encryption pairs, send public key to each party;
\FOR{each iteration j=1,2, ...}
\STATE Randomly sample  $S \subset [N]$
 \FOR{each passive party ($k \ne\ K$) in parallel}
 \STATE $\boldsymbol k$ computes, encrypts and sends $\{[[H_i^k]]\}_{i \in S}$ to party $K$;
 \ENDFOR
 \STATE active party $K$ computes and sends $\{[[\frac{\partial \ell}{\partial H_i}]]\}_{i \in S}$ to all other parties;
 \FOR{each party k=1,2,\dots,K in parallel}
 \STATE $\boldsymbol k$ computes $[[\nabla_k \ell]]$ with Eq. \eqref{equ:grad_H_k}, sends them with random mask to TTP for decryption; \\
 \STATE $\boldsymbol k$ receive and unmask $\nabla_k \ell$ and update $\theta^{j+1}_k = \theta^{j}_k - \eta \nabla_k \ell$;
 \ENDFOR
\ENDFOR
\label{algo1}
\end{algorithmic}
\end{algorithm}

In this paper, we first perform theoretical analysis on the batch label inference task under black-boxed VFL scenario and show that labels can be fully recovered when batch size is smaller than the dimension of the embeded features of the final fully connected layer. Next we demonstrate that label inference attack can be possible using a gradient inversion technique with only the local batch-averaged gradients information. Experiments on multiple datasets show that the accuracy of the batch label reconstruction attack can achieve over $90\%$ accuracy. Built on the success of the label inference attack, we further show that attacker can directly replace labels through gradient-replacement attack under a black-boxed condition without changing VFL protocols. Finally, we evaluate several defense strategies. In particular, under the assumption that parties do not have prior knowledge of other parties' the local model and data in VFL, we introduce a novel technique termed confusional autoencoder (CoAE), based on autoencoder and entropy regularization, to hide the true labels without any noticeable changes to the original protocol. We demonstrate that our technique can successfully block the label inference and replacement attack without hurting as much main task accuracy as existing methods like differential privacy (DP) and gradient sparsification(GS), making it a practical defense strategy in VFL scenarios.

\section{Related Work}
Federated Learning (FL) was originally designed to construct a machine learning environment based on datasets that are distributed among different users while preventing data leakage at the same time. FL can be divided into two main categories, Horizontal Federated Learning (HFL) and Vertical Federated Learning (VFL), based on whether the data is partitioned by samples or by features among users.\cite{Yang-et-al:2019}. In HFL, gradients or model parameters are communicated instead of raw data for collaborative training. Recently, deep neural networks have been adopted to learn sensitive information from those exposed gradients in HFL \cite{Deepleakage2019,DBLP:journals/corr/abs-2001-02610,Yin_2021_CVPR_batch_gradinversion}. Differential Privacy (DP) \cite{Bagdasaryan2018backdoor,xie2020dba} and gradient sparsification (GS) \cite{DBLP:journals/corr/abs-1712-01887} are techniques commonly adopted to preserve privacy in these scenarios.

VFL frameworks have been developed for various models
including trees~\cite{secureboost}, linear and logistic regression
\cite{Gratton2018,Kikuchi2017PrivacyPreservingML,HuADMM}, and neural networks \cite{liu2018ftl,Hu2019FDMLAC}. Because each sample's features are distributed in multiple parties, sample-specific updates are communicated for gradient calculation during training. Communications of sample-level information can be either white-boxed or encrypted, where
privacy-preserving techniques such as Homomorphic Encryption (HE)
\cite{Rivest1978,Acar:2018:SHE:3236632.3214303} is typically applied to
preserve user-level privacy \cite{Yang-et-al:2019}. Recently, label leakage and protection for the vertical federated learning (VFL) framework is studied \cite{Liu_VFLbackdoor,label_leakage_VFL_2021,fulabel}. Liu\cite{Liu_VFLbackdoor} points out label can be inferred from sign of gradients. Li's study \cite{label_leakage_VFL_2021} exploits the difference in the norms of gradients for positive and negative classes where per-sample gradient information is available. Fu\cite{fulabel} proposed three different label inference attack methods in VFL setting: passive model completion, active model completion and direct label inference attack. In model completion attacks, the attacker needs extra auxiliary labeled data for fine-tuning its local 
model. In our work, we attack a more challenging system, where we assume the communicated messages are also black-boxed (e.g. protected by HE), and only the final local model gradients are accessible to attackers, a scenario that is inline with real-world VFL projects, such as the implementation of FATE\cite{JMLR:FATE}.

In addition, previous studies have shown that FL opens doors to backdoor attacks. A typical example is backdoor attacks with model poisoning. Many recent works \cite{Bagdasaryan2018backdoor,pmlr-v97-bhagoji19a} have shown that FL is vulnerable to model poisoning where the attacker manipulates the model's performance on an attacker-chosen backdoor task while maintaining the performance of the main task. These attacks only apply to HFL where model parameters are exposed. There also exists several works which focus on resolving the backdoor attack and improving the safety level of FL systems. A data augmentation approach \cite{PP_data_augment2021} is recently proposed to defend the gradient-based information reconstruction attacks.  Bagdasaryan\cite{Bagdasaryan2018backdoor} introduced a backdoor attack to federated learning by replacing the global model with a targeted poisoning model and discussed effectiveness of possible defense strategies. Bhagoji\cite{pmlr-v97-bhagoji19a} carried out stealthy model poisoning attack to federated learning by alternatively optimizing for stealth and the adversarial objectives. Xie\cite{xie2020dba} introduced distributed backdoor attacks to federated learning. Sun\cite{Sun2019CanYR} studied backdoor and defense strategies in federated learning and show that norm clipping and “weak” differential privacy mitigate the attacks. However, all the works above consider the HFL scenario.

\begin{figure}[!htb]
\includegraphics[width=0.5\linewidth]{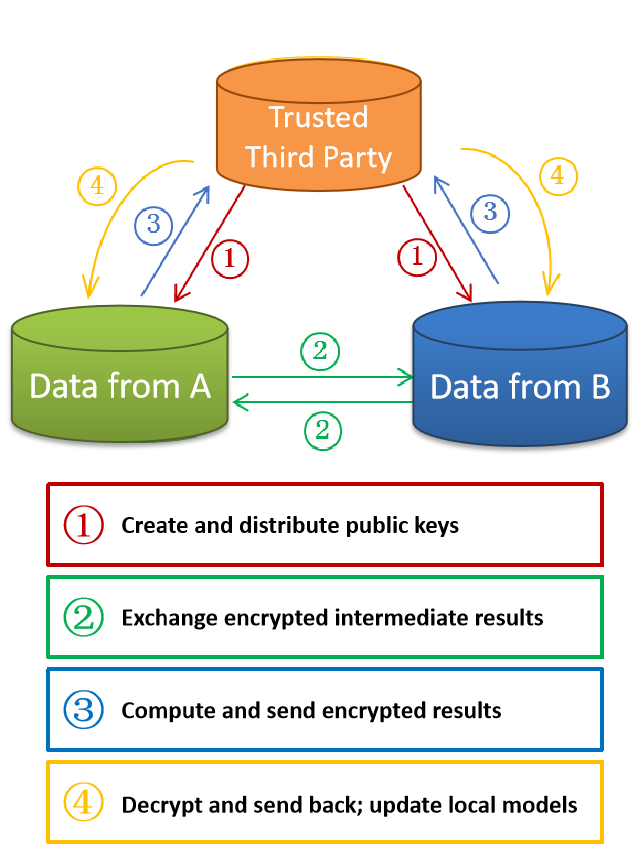}
\centering
\caption{Protocols of a black-boxed VFL system (VFL system with Homomorphic Encryption)}
\label{fig:vfl_blackbox}
\end{figure}

\section{A Black-boxed Vertical Federated Learning Framework with HE Encryption}

In a typical VFL system \cite{Yang-et-al:2019}, $K$ data owners collaboratively train a machine learning model based on a set of data $\{\mathbf{x}_i, y_i\}_{i=1}^{N}$ and only one party has labels $\{y_i\}_{i=1}^{N}$. This is a reasonable assumption in cross-organizational collaborative learning scenarios, because in reality, labels (such as users' credit scores, patients' diagnosis etc.) are expensive to obtain and only exist in one or few of the organizations. Suppose that the feature vector $\mathbf{x}_i$ can be further decomposed into $K$ blocks $\{\mathbf{x}^k_i\}_{k=1}^{K}$, where each block belongs to one owner. 
Without loss of generality, we can assume that the labels are located in party $K$. Then the collaborative training problem can be formulated as 
\begin{align}\label{eq:y_pred}
o_i^K = f(\mathbf{\theta}_1,\dots,\mathbf{\theta}_K; \mathbf{x}_i^1,\dots,\mathbf{x}_i^K)
\end{align}
\begin{align}\label{eq:loss}
\min_{\Theta} \mathcal{L}(\Theta; \mathcal{D})\triangleq \frac{1}{N}\sum^N_{i=1} \ell(o_i^K,y_i^K) + \lambda\sum_{k=1}^K\gamma(\theta_k)
\end{align}
where $\theta_k$ denotes the training parameters of the $k^{th}$ party; $\Theta={[{\mathbf{\theta}_{1}};\dots; {\mathbf{\theta}_{K}}]}$; $N$ denote the total number of training samples; $f(\cdot)$ and $\gamma(\cdot)$ denote the prediction function and regularizer and $\lambda$ is the hyperparameter; Following previous work, we assume each party adopts a sub-model $G_k$ which generates local predictions, i.e. local latent representations $H_i^k$ and the final prediction is made by merging $H_i^k$ with an nonlinear operation $S(\cdot)$, such as softmax function. That is,
\begin{equation}\label{H_k}
H_i^k = G_k(\theta_k, \mathbf{x}_i^k)
\end{equation}
\begin{equation}\label{loss_H_k}
\ell(\mathbf{\theta}_1,\dots,\mathbf{\theta}_K; \mathcal{D}_i)=\ell(S(\sum_{k=1}^K H_i^k),y_i^K)
\end{equation}

where $G_k$ can adopt a wide range of models such as linear and logistic regression, support vector machines, neural networks etc. Let $H_i = \sum_{k=1}^K H_i^k$, then the gradient function has the form
\begin{equation}\label{equ:grad_H_k}
\nabla_k \ell(\mathbf{\theta}_1,\dots,\mathbf{\theta}_K; \mathcal{D}_i) = \frac{\partial \ell}{\partial H_i}\frac{\partial H_i^k}{\partial \theta_k}
\end{equation}

\begin{figure}[!tb]
\includegraphics[width=1\linewidth]{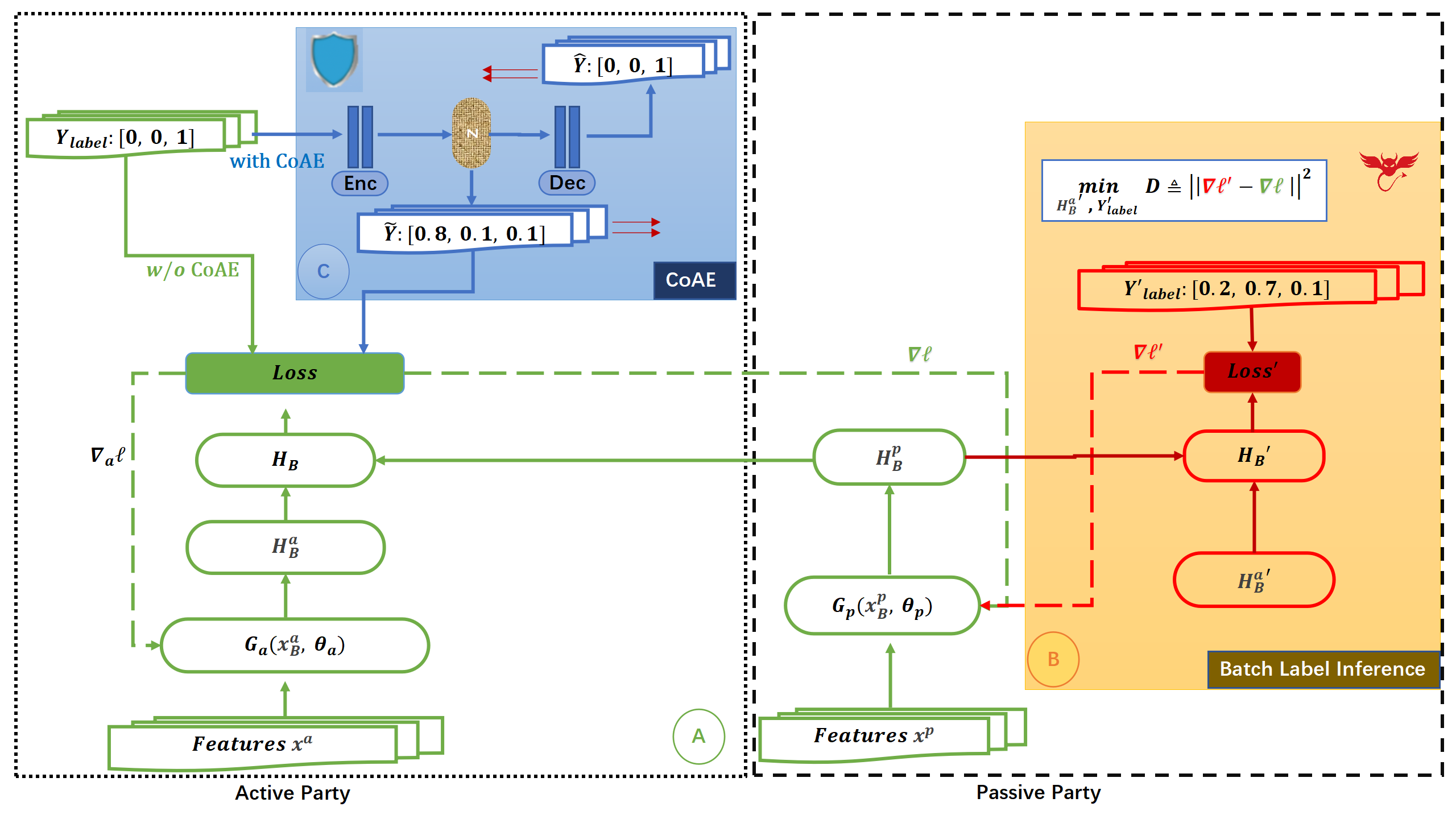}
\centering
\caption{The gradient inversion label inference attack and defense in VFL. ($A$) Basic VFL framework. ($B$) Batch label inference attack (i.e. gradient inversion label inference attack). ($C$) CoAE protection. All the symbols with $B$ as subscripts in this figure denote the batch level arguments. $a,p$ stands for $active$ and $passive$ party respectively.}
\label{fig:vfl_attach_setting}
\end{figure}

We refer the party having the labels the \textit{active party}, and the rest \textit{passive parties}. In a white-box vertical federated learning protocol, each passive party sends \{$H_i^k$\} to active party, and the active party calculates $\{\frac{\partial \ell}{\partial H_i}\}$ and sends it back to passive parties for gradient update. To further protect sample-level information leakage from intermediate results, Homomorphic Encryption(HE), denoted as $[[\cdot]]$, are applied to each sample's communicated updates, \{$H_i^k$\}, and the gradient calculations are performed under encryption. The computed encrypted gradients are then sent to a Trusted Third Party (TTP) for decryption. Specifically, as Figure\ref{fig:vfl_blackbox} illustrated, the following steps are taken. 1) TTP creates encryption pairs, sends public key to each party; 2) Parties encrypt and communicate the intermediate results for gradient calculations;
3) Parties compute encrypted gradients and add additional mask, respectively, and send encrypted values to TTP; 4) TTP decrypts and send the decrypted gradients back to each party; 5) Each party unmask the gradients, update the model parameters accordingly. Note the additional mask is applied to ensure TTP server does not learn gradients from parties, see \cite{Yang-et-al:2019}. This process ensures that only local batch-averaged gradients rather than sample-level gradients are available to the passive party. See also Algorithm \ref{algo1} and Figure \ref{fig:vfl_attach_setting}($A$).

\section{Batch Label Inference Attack} 

\begin{small}
\begin{table}[htb]
\centering
\begin{tabular}{m{2.2cm}<{\centering}|m{1.4cm}<{\centering}m{2.1cm}<{\centering}m{1.2cm}<{\centering}}
\toprule[1.2pt]
{Attack} & {\shortstack{Require\\sample-level\\gradients}} & {\shortstack{Require\\auxiliary\\labeled samples}} & {\shortstack{Number\\of classes\\supported}}\\
\midrule[1.3pt]
{\shortstack{Norm-based\\Scoring\cite{label_leakage_VFL_2021}}} & {\shortstack{\Checkmark}} & {\shortstack{\XSolid }} & {\shortstack{$= 2$}}\\
\specialrule{0em}{2pt}{0pt}
\specialrule{0em}{0pt}{3pt}
{\shortstack{Direction-based\\Scoring\cite{label_leakage_VFL_2021}}} & {\shortstack{\Checkmark}} & {\shortstack{\XSolid }} & {\shortstack{$= 2$}}\\
\specialrule{0em}{2pt}{0pt}
\hline
\specialrule{0em}{0pt}{3pt}
{\shortstack{Passive Model\\Completion\cite{fulabel}}} & {\shortstack{\XSolid }} & {\shortstack{\Checkmark}} & {\shortstack{$\geq 2$}}\\
\specialrule{0em}{2pt}{0pt}
\specialrule{0em}{0pt}{3pt}
{\shortstack{Active Model\\Completion\cite{fulabel}}} & {\shortstack{\XSolid }} & {\shortstack{\Checkmark }} & {\shortstack{$\geq 2$}}\\ 
\specialrule{0em}{2pt}{0pt}
\specialrule{0em}{0pt}{3pt}
{\shortstack{Direct Label\\Inference\cite{fulabel}}} &  {\shortstack{\Checkmark}} & {\shortstack{\XSolid }} & {\shortstack{$\geq 2$}}\\ 
\specialrule{0em}{2pt}{0pt}
\hline
\specialrule{0em}{0pt}{3pt}
{\shortstack{Gradient\\Inversion\\(this work)}} &  {\shortstack{\XSolid }} & {\shortstack{ \XSolid }} & {\shortstack{$\geq 2$}}\\
\bottomrule[1.2pt]
\end{tabular}
\caption{Comparison of different label inference Attacks in VFL setting.}
\label{tab:label_inference_comparison_three_methods}
\end{table}
\end{small}

\subsection{Threat Model} 
Based on Algorithm\ref{algo1}, we consider the following attack model: (i) The attacker has access to the training data of one passive party and it controls both the training procedure and the local model of that passive party; (ii) The attacker only receives batched-averaged local gradients in plain text, but does not receive encrypted per-sample intermediate results; (iii) The attacker is passive and does not modify local updates such as training weights and gradients in the VFL protocol; (iv) The attacker does not control any benign party's training nor does the attacker control the global communication and aggregation protocols. The differences between our model and the VFL framework considered in \cite{label_leakage_VFL_2021, fulabel} is that our threat model further assumes that the communicated messages between parties are private and encrypted, so that per-sample gradient-related information are not available for inference, therefore our attack is black-boxed, and is on the \textit{batch} or \textit{population} level. In addition, we do not assume attacker has auxiliary balanced label samples, which can be difficult or impossible to obtain in real-world applications. We compare our approach with other recent works in detail in Table\ref{tab:label_inference_comparison_three_methods}. In this table, norm-based and direction-based scoring attacks\cite{label_leakage_VFL_2021} can be used for binary classifier task only by utilizing the norm or direction of per-sample gradients to separate the training data into two classes and "Model Completion" denotes the method which is based on adding and fine-tuning a "completion layer" at the passive party with balanced auxiliary data to infer active party's private label. All these methods are designed for label inference in VFL setting.


\begin{algorithm} 
\caption{Black-box batch label inference attack in vertical federated learning}
\textbf{Input}: learning rate $\eta$; Batch $\mathcal{B}$\\
\textbf{Output}: the label recovered from the gradients $\{y_i^{\prime}\}_{i \in \mathcal{B}}$

\begin{algorithmic}[1] 
\STATE \textbf{Passive party do:} 
\STATE receive decrypted local gradients $\nabla\ell$; 
\FOR{$i$ in $\mathcal{B}$}
\STATE Initialize ${H_i^a}^{\prime} \leftarrow \mathcal{N}(0,1), {y_i}^{\prime} \leftarrow \mathcal{N}(0,1)$
\ENDFOR
\FOR{$j \leftarrow 1$ \textbf{to} $iterations$}
\STATE compute local recovery loss with equation (\ref{eq:label_leakage_objective}); \\
\FOR{$i$ in $\mathcal{B}$}
\STATE ${y_i}^{\prime} \leftarrow {y_i}^{\prime}-\eta  \cdot \partial D / \partial {y_i}^{\prime}, {H_i^a}^{\prime} \leftarrow {H_i^a}^{\prime}-\eta \cdot \partial D / \partial {H_i^a}^{\prime}$
\ENDFOR

\ENDFOR
\STATE return $\{y_i^{\prime}\}_{i \in \mathcal{B}}$ 
\label{label_leakage_algorithm}
\end{algorithmic}
\end{algorithm}

\subsection{Methodology} \label{section:ATTACK}
In this section, we first perform analysis on the computed batch-averaged local gradients and its connections with the labels and then show how the success of label inference attack depends on batch size. Next we introduce our attack method, which is based on gradient inversion. 

\begin{lemma}{\bf Direct Label Inference.}\label{lemma_1} In Eq.(\ref{loss_H_k}), if $f$ represents a softmax function and $\ell$ represents a cross-entropy loss, then inferring $\frac{\partial \ell}{\partial H_i}$ is equivalent to inferring labels.
\end{lemma}
\emph{Proof.} $\frac{\partial \ell}{\partial H_i}$ is a $N$-dimension vector where $N$ is the number of classes with the $j^{th}$ element being:
\begin{equation}\label{equ:one_negative}
    \frac{\partial \ell}{\partial H_{i,j}}=\begin{cases}
			S_j, &j\neq y\\ 
			S_j-1, &j=y
			\end{cases}
\end{equation}
where $S_j$ is the softmax function $S_j= \frac{e^{h_j}}{\sum_v e^{h_v}}$ over $H_i$. Here we abuse the notation $y$ to denote the index of the true label. If $\frac{\partial \ell}{\partial H_i}$ is revealed, the label information is known because the $y^{th}$ element of $\frac{\partial \ell}{\partial H_i}$ will have opposite sign compared to others. Therefore we establish that if we know $\frac{\partial \ell}{\partial H_i}$, then we know the exact labels.

As Eq.(\ref{equ:grad_H_k}) shows, for party $k$ in VFL system, 
$H_i^k$ is the logits of the local model of party $k$ for sample $i$ and is a $C$-dimention vector ($C$ is the number of classes for the classification task). 

Let $B$ denotes the size of a batch $\mathcal{B}$. Then its total loss is  ${\ell}_\mathcal{B} =\frac{1}{B} \sum_{i=1}^{B} {\ell}_i$.
Denote the parameters of last fully connected layer $o$ on client $k$ by $\theta_{k,o}$, which is a $C \times M$ matrix, where $M$ is the dimension of the feature embedding. Then the local gradients at the last layer are: 
\begin{equation}\label{equation:batch_gradients_last_layer}
\nabla_{{\theta}_{k,o}} {\ell_\mathcal{B}} = \frac{\partial {\ell_\mathcal{B}}}{\partial H_\mathcal{B}}\frac{\partial H_\mathcal{B}^k}{\partial \theta_{k,o}} 
\end{equation}

where $\nabla_{{\theta}_{k,o}} {\ell_\mathcal{B}} \in \mathbb{R}^{C \times M}$, $\frac{\partial {\ell}_\mathcal{B}}{\partial H_{B}} \in \mathbb{R}^{C \times B}$ and $\frac{\partial H_{\mathcal{B}}^k}{\partial \theta_{k,o}} \in \mathbb{R}^{B \times M}$. 

\begin{theorem}{\bf Denote $\nabla_{{\theta}_{k,o}} {\ell_\mathcal{B}} = \frac{\partial {\ell_\mathcal{B}}}{\partial H_\mathcal{B}}\frac{\partial H_\mathcal{B}^k}{\partial \theta_{k,o}}$ by $Q=UA$, where $Q=\nabla_{{\theta}_{k,o}} {\ell_\mathcal{B}}$ is the matrix of averaged gradients transformed to party $k$, $U=\frac{\partial {\ell}_\mathcal{B}}{\partial H_{B}}$ is the matrix of the unknown and $A=\frac{\partial H_{\mathcal{B}}^k}{\partial \theta_{k,o}}$ is the matrix of local computed back-propagation gradients. When $B \leq rank(A) \leq M$, labels can be 100\% fully recovered.} \label{main_theorem}
\end{theorem}
\emph{Proof.} Because $\frac{\partial H_{\mathcal{B}}^k}{\partial \theta_{k,o}}$ is locally computed, and $\nabla_{{\theta}_{k,o}} {\ell_\mathcal{B}}$ is the averaged gradients thus available to party $k$, so the unknowns are $\frac{\partial {\ell}_\mathcal{B}}{\partial H_{B}}$. Therefore the problem is reduced to solving the linear equations described in Eq.(\ref{equation:batch_gradients_last_layer}), where there are ${C \times B}$ unknown variables and ${C \times M}$ equations in the linear equations. 
Then as $Q^T=A^TU^T$, the matrix form of it is

\begin{small}
\begin{equation}\label{equ:matrix_multiplication}
\left[ \begin{array}{ccc}
q_{11} & \ldots & q_{C1} \\
\vdots & \ddots & \vdots \\
q_{1M} & \ldots & q_{CM}
\end{array} \right] = 
\left[ \begin{array}{ccc}
a_{11} & \ldots & a_{B1} \\
\vdots & \ddots & \vdots \\
a_{1M} & \ldots & a_{BM}
\end{array} \right]
\left[ \begin{array}{ccc}
u_{11} & \ldots & u_{C1} \\
\vdots & \ddots & \vdots \\
u_{1B} & \ldots & u_{CB}
\end{array} \right]
\end{equation}
\end{small}
where $\{u_{cb}\}_{c=1,2,..,C, b=1,2,..,B}$ are unknown arguments.

Eq.(\ref{equ:matrix_multiplication}) can be separated into $C$ parts that are foreign from one another
\begin{small}
\begin{equation}\label{equ:small_linear_simultaneous_equations}
\left[ \begin{array}{ccc}
a_{11} & \ldots & a_{B1}\\
\vdots & \ddots & \vdots\\
a_{1M} & \ldots & a_{BM}
\end{array} \right]
\left[ \begin{array}{c}
u_{c1} \\ \vdots \\ u_{cB}
\end{array} \right]  = 
\left[ \begin{array}{c}
q_{c1}\\ \vdots \\ q_{cM}
\end{array} \right],\ c=1,2,...,C
\end{equation}
\end{small}
In Eq.(\ref{equ:small_linear_simultaneous_equations}), there are $M$ equations and $B$ unknown variables. Therefore, when $B \leq rank(A) \leq M$, all the $C$ groups of Eq.(\ref{equ:small_linear_simultaneous_equations}) has only one group of solution. According to Lemma \ref{lemma_1}, the recovery rate the label should be $1.0$. That is, when the batch size is no larger than the size of embedded features, labels are leaked with 100\%.  

When $B > rank(A)$, there are more than one solution. However, additional constraints can be exploited. First, each column in $\frac{\partial {\ell}_\mathcal{B}}{\partial H_{B}}$ represents a sample $i$'s gradient with respect to $H_i$, and it should have only one element that is less than zero and all the other elements should be in the range of $[0,1]$. From Eq.(\ref{equ:one_negative}), the sum of the elements in each column should be zero. What's more, there is no need to find the exact value of each element of unknown matrix. Instead, the attacker only needs to find out which element is negative. In this case, multiple solutions may lead to the same conclusion about the labels. 

Inspired by the attack performed by \cite{Deepleakage2019}, we adopt a deep learning approach to accomplish this batch-level label inference attack, see Figure \ref{fig:vfl_attach_setting}($A$, $B$) and Algorithm \ref{label_leakage_algorithm}. Specifically, the passive party (attacker) $p$ sets up an internal model which tries to guess the labels and communicated intermediate results $\{H_i^a\}_{i \in \mathcal{B}}$ for every sample $i$ in a batch $\mathcal{B}$ so that the simulated local gradients match the observed ones. First, for each sample $i$ in each batch $\mathcal{B}$ it randomly initializes $y_i'$ and ${H_i^a}'$, then computes its gradients using:
\begin{equation}\label{eq:label_leakage_gradient}
 \nabla \ell^{\prime} = \sum_{i \in \mathcal{B}}\partial \ell\left(\operatorname{softmax}(H_i^p + {H_i^a}^{\prime}), {y_i}^{\prime}\right) / \partial \theta_{p}
\end{equation}
Then the final objective is to match the observed and simulated gradients:
\begin{equation}\label{eq:label_leakage_objective}
\min_{H_a',y'} \mathcal{D}  \triangleq
\left\|\nabla \ell^{\prime}-\nabla \ell\right\|^{2}
\end{equation}

We term our attack \textbf{gradient inversion} and implement the optimization with gradient descend in Algorithm \ref{label_leakage_algorithm}. We show experimentally that this reconstruction can be quite powerful achieving an over $90\%$ label reconstruction rate for large batch size. Therefore, we conclude that batch averaging is not sufficient to protect the label information in Black-boxed VFL setting. In Section\ref{section:DEFENCE}, we propose a novel defense technique to protect labels.


\section{Label Replacement Attacks}
In previous sections, we focus on label inference as the adversarial objective. In this section, we further consider how the adversary is going to leverage the label information for its own benefit in the VFL framework. Specifically, we consider a backdoor attack where an adversary replace a small amount of labels with the target labels, which is commonly studied in the HFL scenarios\cite{Bagdasaryan2018backdoor,pmlr-v97-bhagoji19a,xie2020dba}

\begin{figure}[tb]
\includegraphics[width=1\linewidth]{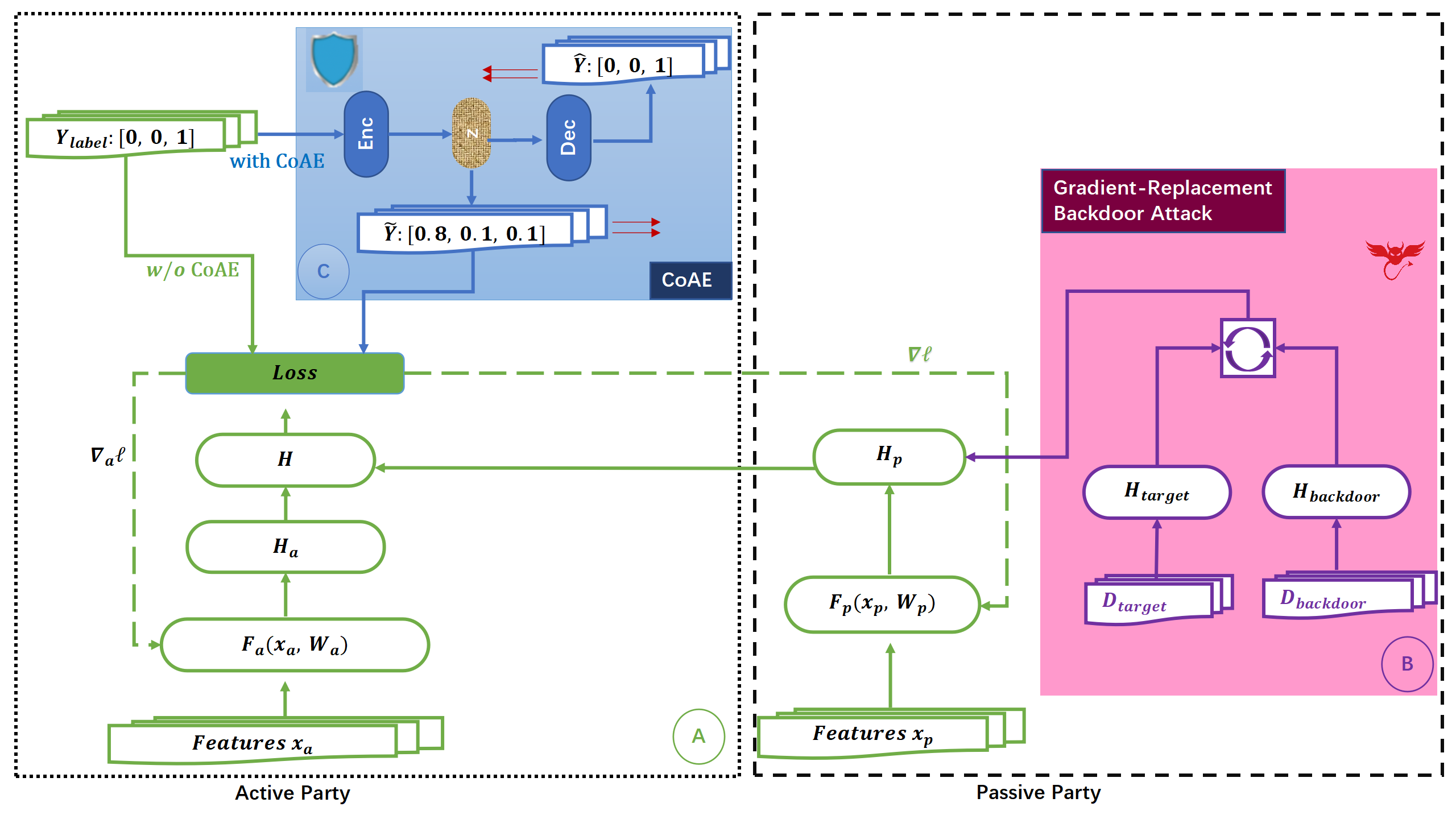}
\centering
\caption{The label replacement backdoor attack and defense in VFL. ($A$) Basic VFL framework. ($B$) Label replacement backdoor. ($C$) CoAE protection.}

\label{fig:vfl_attack_backdoor_setting}
\end{figure}

\begin{figure}[tb]
\centering
\includegraphics[width=1\linewidth]{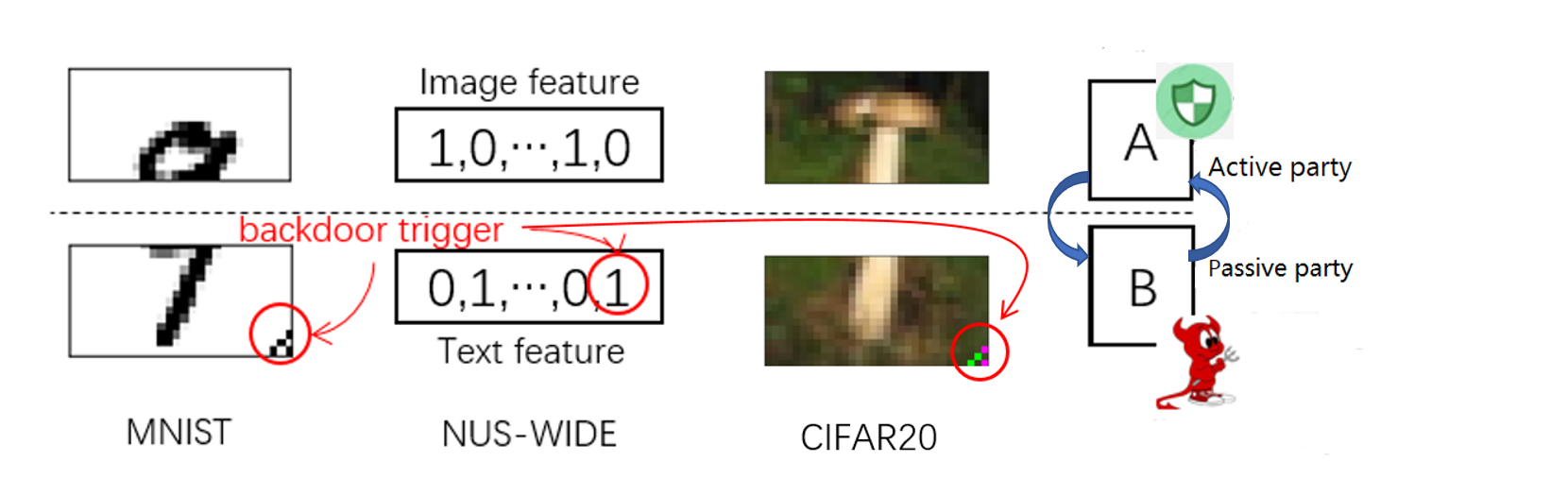}
\centering
\caption{Backdoor task settings}
\label{fig:backdoor_task_setting}
\end{figure}

In our setting, the backdoor task is to assign an attacker-chosen label to input data with a specific pattern (i.e. a trigger), as shown in Figure \ref{fig:backdoor_task_setting}. Also, the attacker aims to train a model which achieves high performance on both the original task and the backdoor task. The adversarial objective can be written as:
\begin{equation}\label{eq:problem}
\begin{split}
\min_{\Theta} \mathcal{L}^b(\Theta; \mathcal{D}) \triangleq &\frac{1}{N_{cln}}\sum_{i \in \mathcal{D}_{cln}} \ell(o_i^K,y_i^K) + \\
&\frac{1}{N_{poi}}\sum_{i \in \mathcal{D}_{poi}} \ell(o_i^K,r_i^K) 
\end{split}
\end{equation}
where $o_i^K$ is defined in equation \ref{eq:y_pred},  $\mathcal{D}_{cln}$,$N_{cln}$,$\mathcal{D}_{poi}$,$N_{poi}$ denote the clean dataset, number of samples in clean dataset, poisoned dataset and number of samples in poisoned dataset, respectively. $r_i^K$ denotes the target label. 
From algorithm \ref{algo1}, the only information the passive parties obtain is the batch-averaged gradients. Therefore the main challenge is for the passive parties to assign targeted labels to specific input data with trigger features \textit{without direct access to labels}. 

This backdoor task is built on the label inference attack discussed previously since once a passive party learns the corresponding true labels of the each sample, it can carefully design the encrypted communicated messages to replace thae true label with a target label. Following Eq.(\ref{equ:one_negative}), after label inference, $\frac{\partial \ell}{\partial H_i}$ is obtained and the $y^{th}$ element of $\frac{\partial \ell}{\partial H_i}$ has opposite sign as compared to others.

To inject the backdoor attack, the passive party only needs to replace $\frac{\partial \ell}{\partial H_i}$ with  
\begin{equation}
    \left(\frac{\partial \ell}{\partial H_{i,j}}\right)^{b}=\begin{cases}
			S_j, &j\neq \tau\\ 
			S_j-1, &j=\tau
			\end{cases}
\end{equation}
Where $\tau$ is the target label. Note that under the black-boxed VFL scenario, this injection should be and can be performed under HE encryption without knowing the plain-text value of $S_j$ due to the additive property of HE, that is, 
\begin{equation}
    \left[\left[\left(\frac{\partial \ell}{\partial H_{i,j}}\right)^{b}\right]\right]=\begin{cases}
			\left[\left[S_j\right]\right] , &j\neq \tau, j\neq y \\ 
			\left[\left[S_j\right]\right] - [[1]], &j=\tau \\
			\left[\left[S_j\right]\right] + [[1]], &j= y \\
			\end{cases}
\end{equation}
Therefore we can exploit the recovered labels using label inference attack to perform backdoor attacks.


\begin{algorithm}[!htb]
\caption{Label replacement backdoor}  

\textbf{Input}: \\
\hspace*{0.4cm} $D_{target}$: Clean target dataset; \\
\hspace*{0.4cm} $D_{backdoor}$: poisoned dataset; \\ 
\hspace*{0.4cm} $\mathcal{X}$: current batch of training dataset; \\ 
\hspace*{0.4cm} $\gamma$: amplify rate

\begin{algorithmic}[1] 
\STATE \textbf{Forward Propogation:} 
\STATE Passive party computes $\{H_i\}_{i \in \mathcal{X}}$; 
\STATE empty pair set $\mathcal{P}$;
\FOR{$x_i$ in $\mathcal{X}$} 
\IF{$x_i$ belongs to $\mathcal{D}_{backdoor}$} 
\STATE random select $j$ from $D_{target}$;
\STATE $\mathcal{P} += <i,j> $; 
\STATE replace $H_j$ with $H_i$;
\ENDIF
\ENDFOR
\STATE Passive party encrypts and sends $[[H_{\mathcal{X}}]]$ to active party; 
\STATE \textbf{Backward Propagation:} 
\FOR{$<i,j>$ in $\mathcal{P}$}
\STATE replace $[[\frac{\partial \ell}{\partial H_i}]]$ with $\gamma[[\frac{\partial \ell}{\partial H_j}]]$; 
\ENDFOR
\label{algo_gradient_poison_1}
\end{algorithmic}
\end{algorithm}

Such a backdoor attack requires the label inference step as a prerequisite, which requires running and observing the VFL process separately first. 
In fact, for this backdoor attack, we can also eliminate the label inference step completely and directly replace labels via replacing encrypted gradients. We assume that the passive attacker knows a few clean samples with the targeted label, marked as $\mathcal{D}_{target}$. Note that, these clean samples are from the original local dataset of the attacker and as few as 1 clean sample from target class is needed. With this assumption, we propose to inject a gradient-replacement backdoor to the learning process directly as follows: 
\begin{itemize}
    \item In the forward propagation, the attacker performs local computations of $H_i^k$, but for each poisoned sample $i$, it randomly selects one sample $j$ from $\mathcal{D}_{target}$, replaces encrypted intermediate results $[[H_j^k]]_{j \in \mathcal{D}_{target}}$ with $[[H_i^k]]_{i \in \mathcal{D}_{backdoor}}$ and records the pair $<i,j>$;
     \item In the backward propagation, for each pair $<i,j>$, the attacker replaces the encrypted intermediate gradients of sample $i$ it receives with encrypted gradients that exits at the position for sample $j$, update the model parameters using $\gamma[[\frac{\partial \ell}{\partial H_j}]]\frac{\partial H_i^k}{\partial \theta_k}$ according to Eq.(\ref{equ:grad_H_k}) where $\frac{\partial \ell}{\partial H_i}$ is replaced by $\frac{\partial \ell}{\partial H_j}$; 
\end{itemize}
Here $\gamma$ is an amplify rate that we adjust to control the level of backdoor. This can also be understood as an identity stealth strategy where the backdoor sample steals a target's identity. By using such a strategy, the passive party will obtain the corresponding gradients with respect to the targeted label instead of its own therefore its local backdoor updates will be successful. To further prevent the active party from learning a reasonable mapping of the labels and intermediate results $H_i$ of the poisoned samples, the passive party can instead output a random-valued vector to the active party for each poisoned sample during training. See Algorithm \ref{algo_gradient_poison_1} and Figure \ref{fig:vfl_attack_backdoor_setting}($A$, $B$) for full details.

\begin{algorithm}
\caption{Label Disguise via training CoAE} 
\label{autoencoder_train}
\textbf{Input}: \\
\hspace*{0.4cm} the learning rate $\eta$;\\
\hspace*{0.4cm} number of classes $\boldsymbol{c}$, batch size $\boldsymbol{N}$;\\
\hspace*{0.4cm} the maximum number of epochs $\boldsymbol{E}$;\\ 
\textbf{Output}: trained $W_e$, $W_d$
\begin{algorithmic}[1] 
\FOR{$i \leftarrow 1$ to $E$ } 
\STATE Randomly generate one-hot labels: $y \in R^{\mathrm{N} \times \mathrm{c}}$\\
\STATE Generate fake and reconstructed labels by Eq. (\ref{encoder});
\STATE Compute total loss $L$ by Eq. (\ref{loss_total}) and update parameters: \\
$W_e \leftarrow  W_e - \eta \cdot \partial L / \partial W_e$\\ $W_d \leftarrow  W_d - \eta \cdot \partial L / \partial W_d$
\ENDFOR 
\STATE \textbf{return} $W_e$ and $W_d$ 
\end{algorithmic} 
\end{algorithm}

\section{Defense} \label{section:DEFENCE}

In the previous sections, we have shown that label leakage is a challenging problems to VFL. In this section, we discuss possible routes for protection. Especially, the active party needs to establish a stronger defense mechanism. The common defense mechanisms previously adopted include differential privacy and gradient sparsification. However these techniques usually suffer from accuracy loss, which might be unacceptable in certain scenarios. Another line of work is to exploit the distribution differences of per-sample gradient values \cite{label_leakage_VFL_2021}, but they are not always accessible. 

Here we propose a simple yet effective label disguise technique where the active party learns to transform the original labels to a set of "soft fake labels" so that 1) these soft fake labels contrast with the original ones, i.e. the fake labels and the original ones are likely not the same after the transformation; 2) the original labels can be reconstructed almost losslessly; 3) the fake labels should introduce as much confusion as possible to prevent the passive party to infer the true labels. For example, a simple mapping function (e.g., a function that assigns label "dog" to "cat" and "cat" to "dog") would satisfy the first two conditions but not the last one, since if a party knows the true label of one sample, it would know the true labels of all samples belonging to the same class. To increase confusion, we propose learning a confusional autoencoder (CoAE) to establish a mapping such that one label will be transformed into a soft label with higher probability for each alternative class. For example, a dog is mapped to [0.5,0.5] probability of dog and cat, respectively. Whereas autoencoder is simple and effective for hiding true labels, confusion is important in changing the probability distribution of classes, making label leakage attack harder to succeed.  

\begin{algorithm} 
\caption{VFL with CoAE protection} 
\label{algo:coAE} 
\textbf{Input}: \\
\hspace*{0.4cm}$\boldsymbol{X}$ :Input data;\\
\hspace*{0.4cm}$\boldsymbol{Y_{label}}$ :Label of $\boldsymbol{X}$ ;\\
\hspace*{0.4cm}$W_e, W_d$: trained encoder and decoder parameters;\\
\hspace*{0.4cm}$\boldsymbol{f(\cdot)}$: differentiable neural network model;\\
\textbf{Procedure}:
\begin{algorithmic}[1] 
\STATE \textbf{Training Procedure}
\STATE Get fake soft label: $\tilde{Y}=\operatorname{Enc}(Y_{label},W_e)$
\STATE Get the predicted soft label from VFL: $Y_p = f(H)$
\STATE Compute cross entropy loss: $\ell^{CE}=CE(\tilde{Y},Y_p)$ \STATE Replace original gradients with $\nabla\ell^{CE}$
and send to passive party;
\STATE \textbf{Inference Procedure}
\STATE Reconstruct true labels : $Y_t =\operatorname{argmax}(\operatorname{Dec}\left(f(H)\right)$
\end{algorithmic} 
\end{algorithm}

Algorithm \ref{autoencoder_train} and Figure \ref{fig:vfl_attach_setting}($C$) describe the training procedure. The encoder takes as input ground-truth labels $y$ and outputs fake labels $\tilde{y}$, while the decoder takes as input fake labels $\tilde{y}$ and reconstruct original true labels $\hat{y}$. That is:
\begin{equation}
\label{encoder}
\begin{split}
 \tilde{y} = Enc({y};W_e)\\
 \hat{y} = Dec(\tilde{y},W_d)\\
\end{split}
\end{equation}
Where $W_e$ and $W_d$ are the parameters for encoder and decoder,respectively. To satisfy our conditions, we introduce a contrastive loss and an entropy loss, respectively:

\begin{equation}
\label{ae_loss}
\begin{split}
L_{contra} = CE(y,\hat{y}) - \lambda_1 CE(y,\tilde{y}) \\
L_{entropy} = Entropy(\tilde{y})\\
\end{split}
\end{equation}

Where $CE(\cdot)$ is the cross-entropy loss. Then, we form the final learning objective function as:
\begin{equation}\label{loss_total}
L = L_{contra} - \lambda_2 L_{entropy} 
\end{equation}
Here $L_{contra}$ is the contrastive loss that enables the CoAE to reconstruct true labels from fake ones while forcing the fake labels to be different from the original labels; $L_{entropy}$ is the entropy loss that maps each true label to multiple alternative labels ("confusion"); $\lambda_s, s \in \{1,2\}$ are loss weights. After trained, the active party can leverage the CoAE to produce fake labels and use these fakes labels to compute gradients in VFL, thereby preventing label leakage attack (Algorithm \ref{algo:coAE}).
When performing inference, the active party transforms the predicted labels back to true labels using decoder. 

The proposed CoAE is also proved to be effective for defending gradient-replacement backdoor attack due to its confusion with soft fake labels, as will shown in Experiments. 


Note that our defense strategy is focusing on preventing direct exploitation of the relationship between label information and local gradients so that attackers can not use sign and values of gradient elements to deduce directly or through deep learning the correct labels. Also, our defense strategy does not hurt the main task's accuracy. It exploits the fact that active party controls its local model and other parties do not have prior knowledge about the active party's data or labels. It aims to create targeted transformation and confusion rather than reduce the amount information transmitted to other parties, as in methods like gradient sparsification, which are in general not effective to label leakage attacks. Therefore the proposed CoAE is not designed for scenarios when attacker has prior auxiliary information about labels and can learn a reasonable mapping between their data and labels. Nevertheless, the proposed defense is orthogonal to other information-reduction techniques and can be used in combine for an universal defense strategy, as demonstrated in experiments. 

\section{Experiments}
In our experiments, we first evaluate the effectiveness of our proposed attacks. Then we evaluate the performance of various defense strategies, including our proposed CoAE defense at the active party in a black-boxed VFL training protocol. 

\subsection{Models and Dataset}
\textbf{NUS-WIDE dataset} In this dataset, each sample has 634 image features and 1000 text features. We partition the data into an active party with image features, and the passive party with text features. In the following experiments for batch label inference attack, all 81 labels are used to study the impact of number of classes. For the following label replacement backdoor attack, the backdoor trigger is the last text feature equals 1. The data samples that satisfy this feature are less that $1\%$ in both training set and testing set, thus it's difficult to detect this backdoor using a validation set. In the following experiments, we choose five labels to form our train and test dataset: ['buildings', 'grass', 'animal', 'water', 'person']. The target label is set to $class_0$ and ten target samples are randomly chosen. A 2-layer MLP model is used for this dataset with 32 neurons in the middle layer.\\
\textbf{MNIST and CIFAR dataset} We use both CIFAR10 and CIFAR100 in our experiments. For these three datasets, We evenly split each image into two halves and assign them to the active and passive party, respectively. All labels of both datasets are used through out the experiment. Following \cite{DBLP:journals/corr/abs-1708-06733}, we inject trigger 255 at pixel positions [25,27], [27,25], [26,26] and [27,27] for the MNIST dataset. We randomly select 600 samples from the 60000 training samples and 100 samples from the 10000 testing samples and mark them with this trigger. As for the CIFAR dataset, pixels [29,31] and [30,30] are set to [0,255,0] along the channels, and pixels [31,29] and [31,31] are set to [255,0,255] (see Figure \ref{fig:backdoor_task_setting}). We randomly mark 100 samples from 10000 training data and 20 samples from 2000 test data with this trigger ($\mathcal{D}_{backdoor}$). The target label is randomly chosen. Then 10 target samples are randomly chosen from that class. A 2-layer MLP model is used for MNIST dataset with 32 neurons in the middle layer. Resnet18 is used as the backbone model for both CIFAR10 and CIFAR100 datasets. When comparing with model completion attack, Resnet20 is used to keep consistency with the original work.


\begin{figure}[htb]
\centering
\subfigure[MNIST]{
\begin{minipage}[t]{0.47\linewidth}
\centering
\includegraphics[width=1\linewidth]{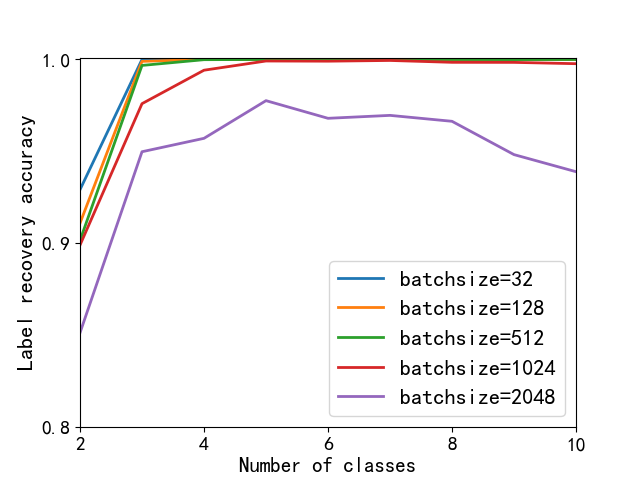}
\end{minipage}
}
\subfigure[NUSWIDE]{
\begin{minipage}[t]{0.47\linewidth}
\centering
\includegraphics[width=1\linewidth]{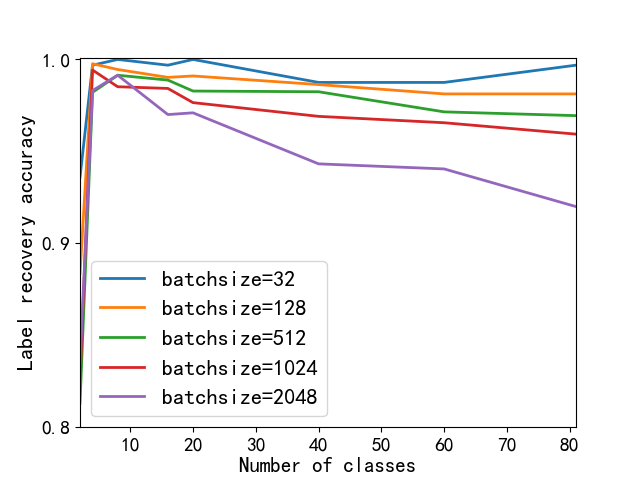}
\end{minipage}
}
\caption{Gradient Inversion-based Label inference accuracy as a function of the number of classes for various batch size.}
\label{fig:label_recovery}
\end{figure}

\subsection{Attacks at Passive Party}
\subsubsection{Batch-level Label Inference Attack}
To test the effectiveness of the label inference attack, we use recovery accuracy as our metric. If the predicted label matches the true label, the recovery is successful for that sample. The recovery accuracy is then the ratio of successfully reconstructed samples. The samples and the classes are randomly selected for each experiment and each run is repeated 30 times. The results are shown in Figure \ref{fig:label_recovery} with different batch size and different number of classes. It can be seen that the batch label inference attack can be quite successfully for various batch size and number of classes, despite the fact that the only clue is the batch-averaged gradients. As the batch size increases, the recovery rate gradually drops, indicating the inference task is more difficult. Another interesting finding is that as the number of classes increases, the recovery rate increases at first, likely due to the fact that the diversity of the labels reduces the possible combinations in the search space. This is consistent with \cite{Yin_2021_CVPR_batch_gradinversion}, where the assumption for successfully guessing the labels in a batch is that the number of classes exceeds the batch size. As the number of classes continues to grow (e.g. to 81 in NUS-WIDE), we also observe a drop in recovery rate, likely due to increasing difficulty for solving the optimization problem.

\begin{small}
\begin{table}[htb]\small
\centering
\begin{tabular}{m{1.8cm}<{\centering}|m{0.9cm}<{\centering}m{0.9cm}<{\centering}m{0.9cm}<{\centering}m{1.8cm}<{\centering}}
\toprule[1.2pt]
{Attack} & {Dataset} & {\shortstack{Number\\of classes}} & {Parameter} & {Attack ACC}\\ 
\midrule[1.3pt]
\multirow{6}*{\shortstack{Passive Model\\Completion\cite{fulabel}}} & \multirow{3}*{CIFAR10} & \multirow{3}*{$10$} & 1 & 0.618\\ 
~ & ~ & ~ & 4 & 0.698\\
~ & ~ & ~ & 12 & 0.715\\
\cline{4-5}
\specialrule{0em}{1pt}{0pt}
~ & \multirow{3}*{CIFAR100} & \multirow{3}*{$100$} & 1 & 0.085 (0.238)\\
~ & ~ & ~ & 4 & 0.181 (0.409)\\
~ & ~ & ~ & 12 & 0.252 (0.487)\\
\specialrule{0em}{1pt}{0pt}
\cline{4-5}
\specialrule{0em}{0pt}{1pt}
\multirow{6}*{\shortstack{Active Model\\Completion\cite{fulabel}}} & \multirow{3}*{CIFAR10} & \multirow{3}*{$10$} & 1 & 0.658\\ 
~ & ~ & ~ & 4 & 0.741\\
~ & ~ & ~ & 12 & 0.748\\
\cline{4-5}
\specialrule{0em}{1pt}{0pt}
~ & \multirow{3}*{CIFAR100} & \multirow{3}*{$100$} & 1 & 0.125 (0.310)\\
~ & ~ & ~ & 4 & 0.261 (0.531)\\
~ & ~ & ~ & 12 & 0.335 (0.594)\\
\specialrule{0em}{1pt}{0pt}
\hline
\specialrule{0em}{0pt}{1pt}
\multirow{6}*{\shortstack{Gradient\\Inversion\\(this work)}} & \multirow{3}*{CIFAR10} & \multirow{3}*{$10$} & 128 & 0.977\\ 
~ & ~ & ~ & 512 & 0.934\\
~ & ~ & ~ & 2048 & 0.893\\
\cline{4-5}
\specialrule{0em}{1pt}{0pt}
~ & \multirow{3}*{CIFAR100} & \multirow{3}*{$100$} & 128 & 0.999\\
~ & ~ & ~ & 512 & 0.969\\
~ & ~ & ~ & 2048 & 0.922\\
\bottomrule[1.2pt]
\end{tabular}
\caption{Comparison of Different Label Inference Attacks in VFL setting. The numbers in parentheses are top-5 accuracy, others are top-1 accuracy. For Passive and Active Model Completion Attack, the parameters are the quantity of auxiliary labeled data for each class whereas for Gradient Inversion Attack, the parameters are the batch size.}
\label{tab:label_inference_comparison}
\end{table}
\end{small}

We also compared our batch-level label reconstruction attack with the three label attacks proposed in \cite{fulabel}, namely, the passive and active model completion and sample-level label inference attack, on attack accuracy and main task accuracy. We strictly follow the settings Fu\cite{fulabel} used in their experiment which means the last layer for aggregating the logits from both parties are slightly different in model completion attacks and direct label inference attack. Note the attacks proposed by Li\cite{label_leakage_VFL_2021} are also sample-level inference attack which are principally similar to \cite{fulabel} but limited to binary classification, so we did not compare with them. Table\ref{tab:label_inference_comparison} shows the results on CIFAR10 and CIFAR100 datasets. For CIFAR10 dataset we report only the top-1 accuracy while for CIFAR100 dataset we report both the top-1 and top-5 accuracy on both the main task and the attack following the settings in \cite{fulabel}. It's clear that our batch-level label inference attack beats both the passive and the active model completion attack with a high attack accuracy with a similar main task accuracy. It's not surprising that direct label inference attack, also sample-level label inference attack, completely recovers all the label since in this case the batch size $B = 1 \leq rank(A)$. As mentioned in section \ref{section:ATTACK}, the attacker should be able to recover each label precisely.

\subsubsection{Label Replacement Backdoor Attack}
\begin{figure}[!tb]
\centering
\subfigure[MNIST]{
\begin{minipage}[t]{0.47\linewidth}
\centering
\includegraphics[width=1\linewidth]{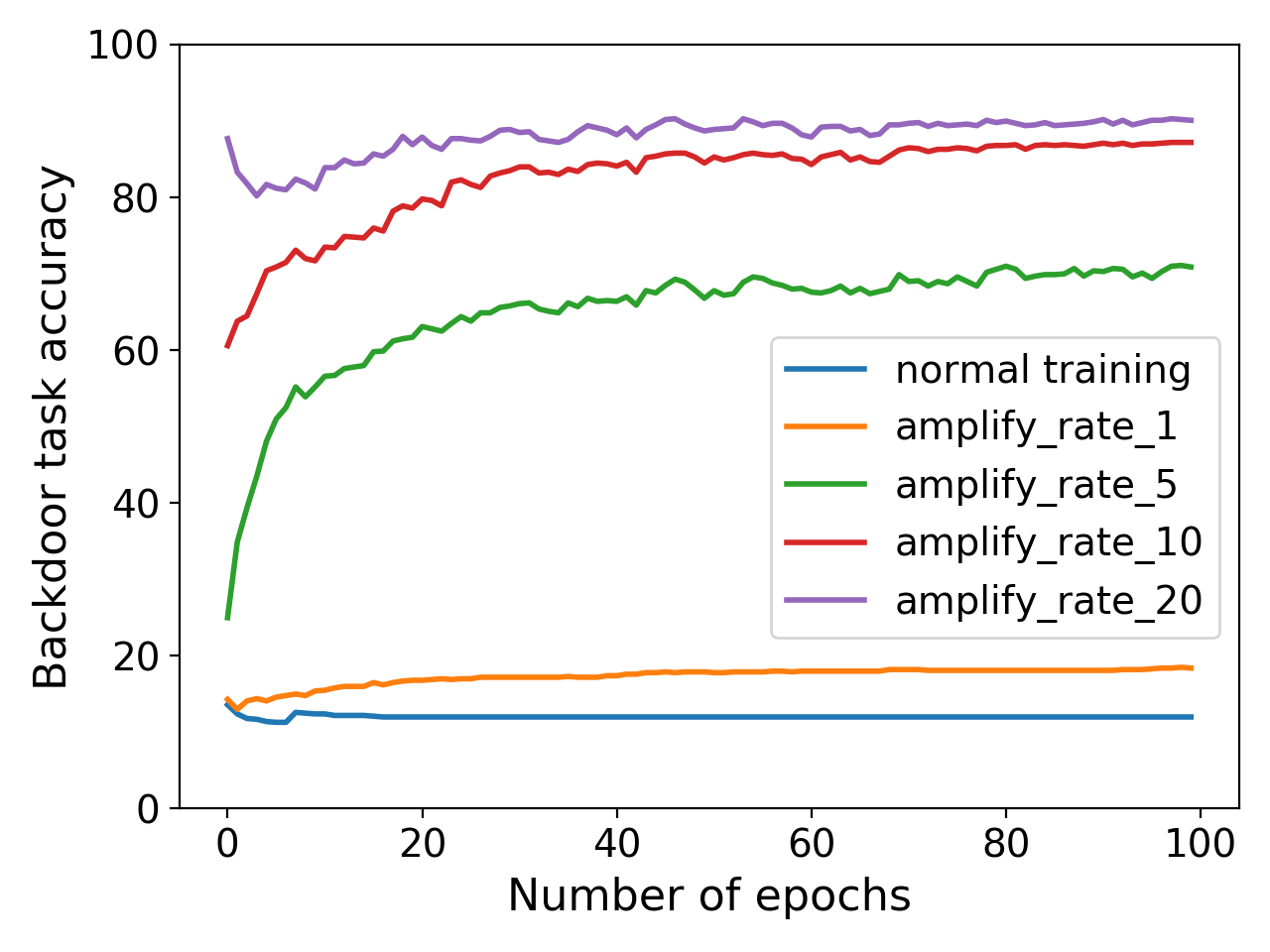}
\end{minipage}
}
\subfigure[MNIST]{
\begin{minipage}[t]{0.47\linewidth}
\centering
\includegraphics[width=1\linewidth]{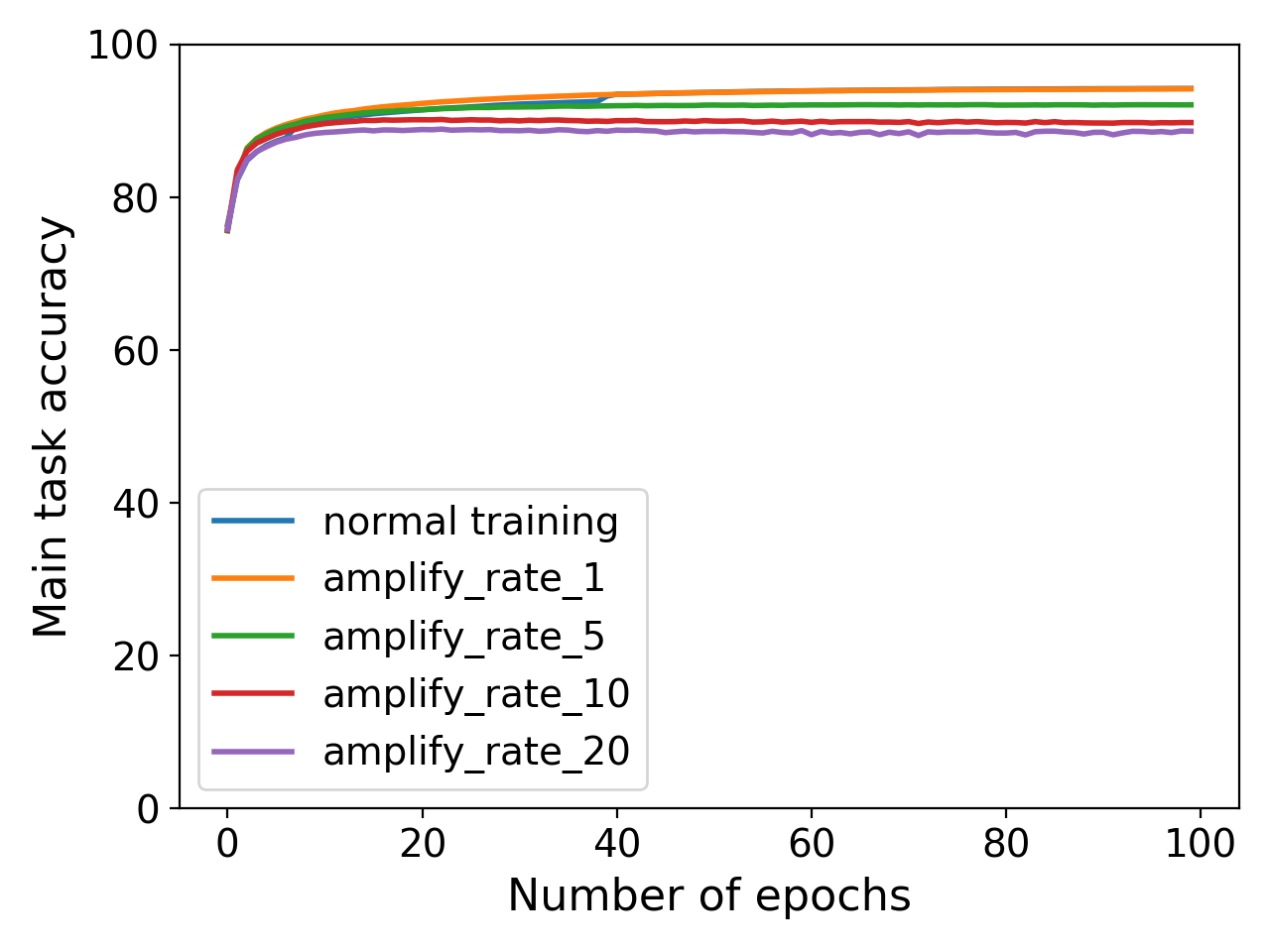}
\end{minipage}
}

\subfigure[NUS-WIDE]{
\begin{minipage}[t]{0.47\linewidth}
\centering
\includegraphics[width=1\linewidth]{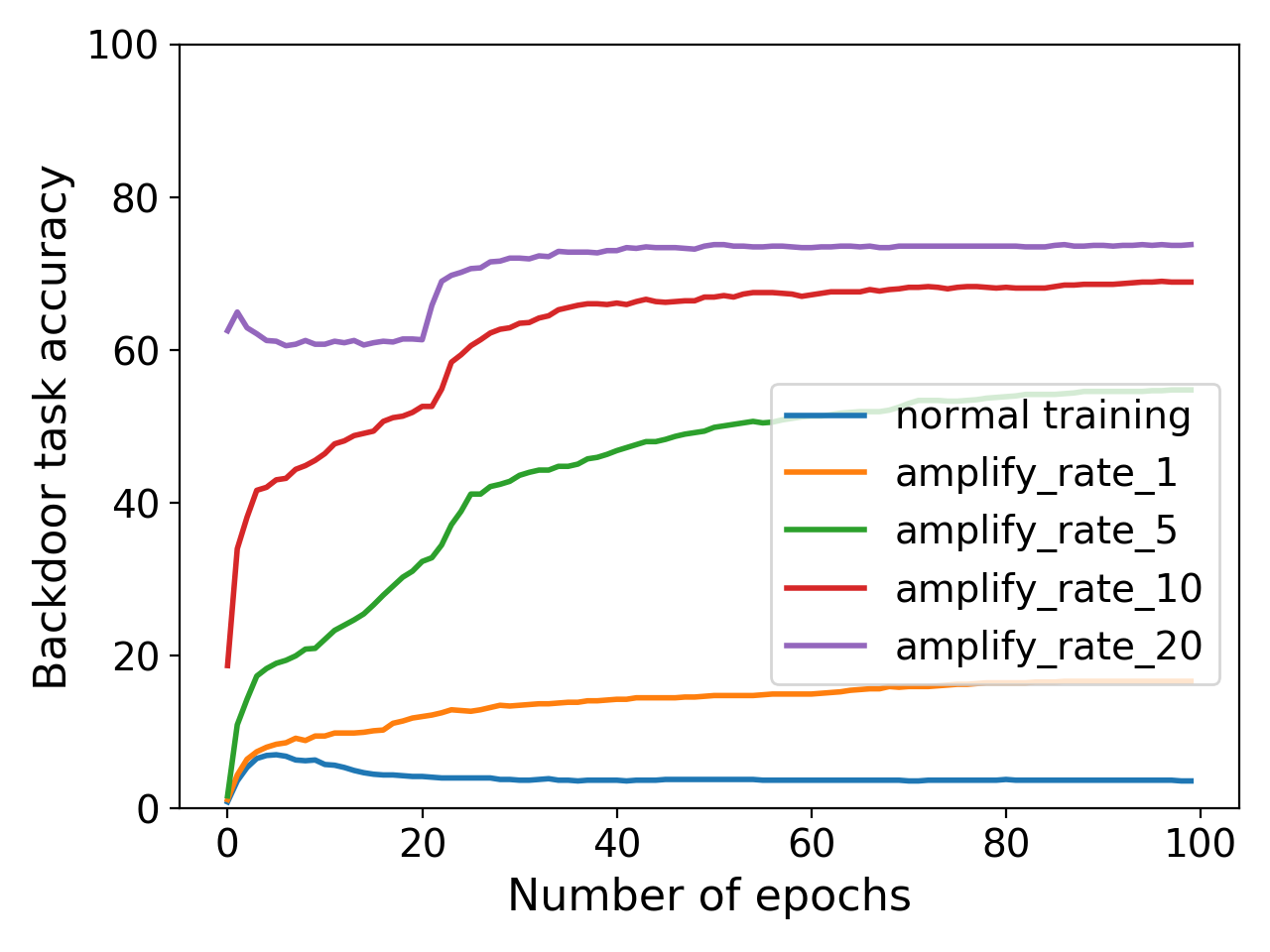}
\end{minipage}
}
\subfigure[NUS-WIDE]{
\begin{minipage}[t]{0.47\linewidth}
\centering
\includegraphics[width=1\linewidth]{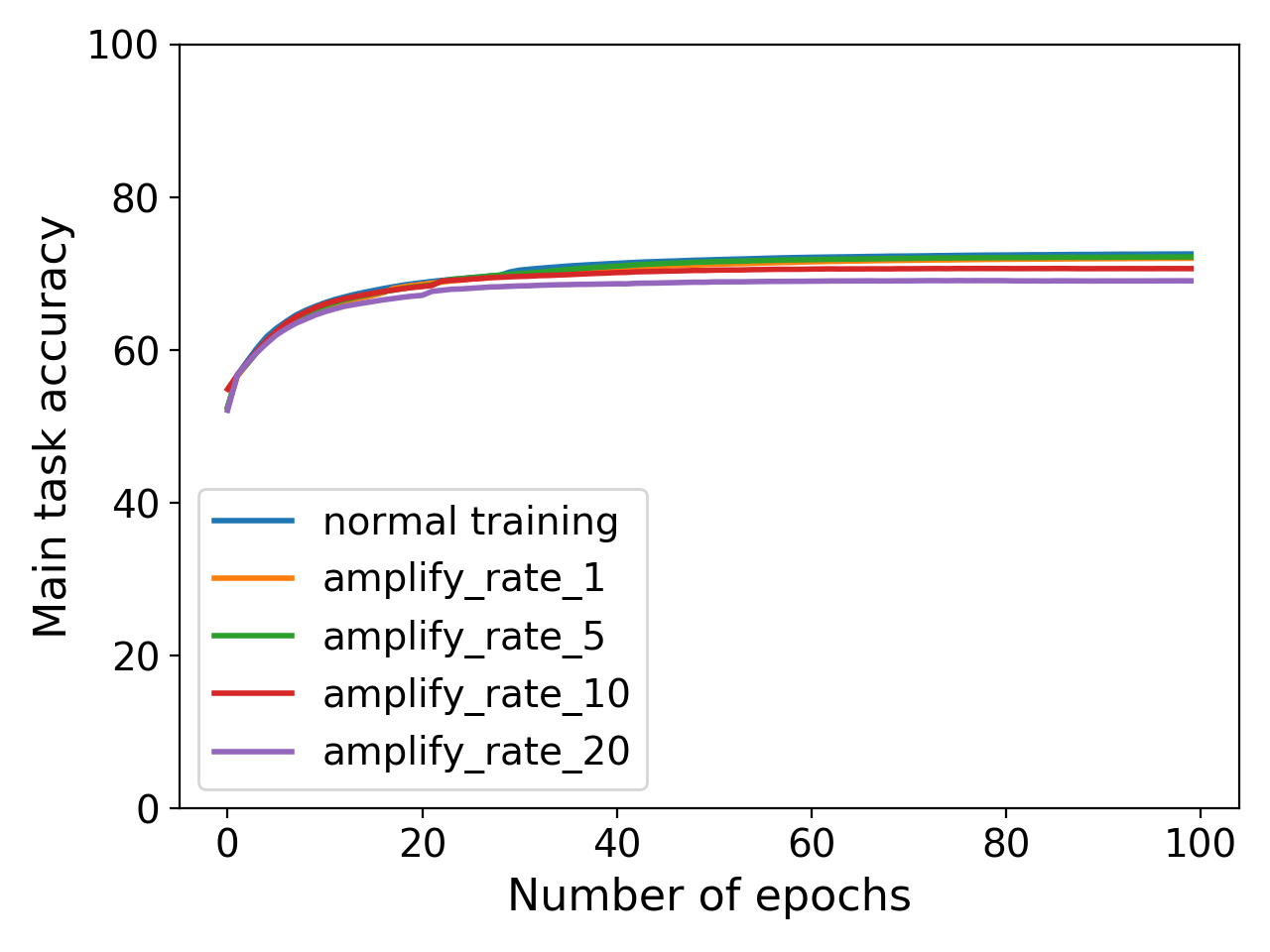}
\end{minipage}
}

\subfigure[CIFAR20]{
\begin{minipage}[t]{0.47\linewidth}
\centering
\includegraphics[width=1\linewidth]{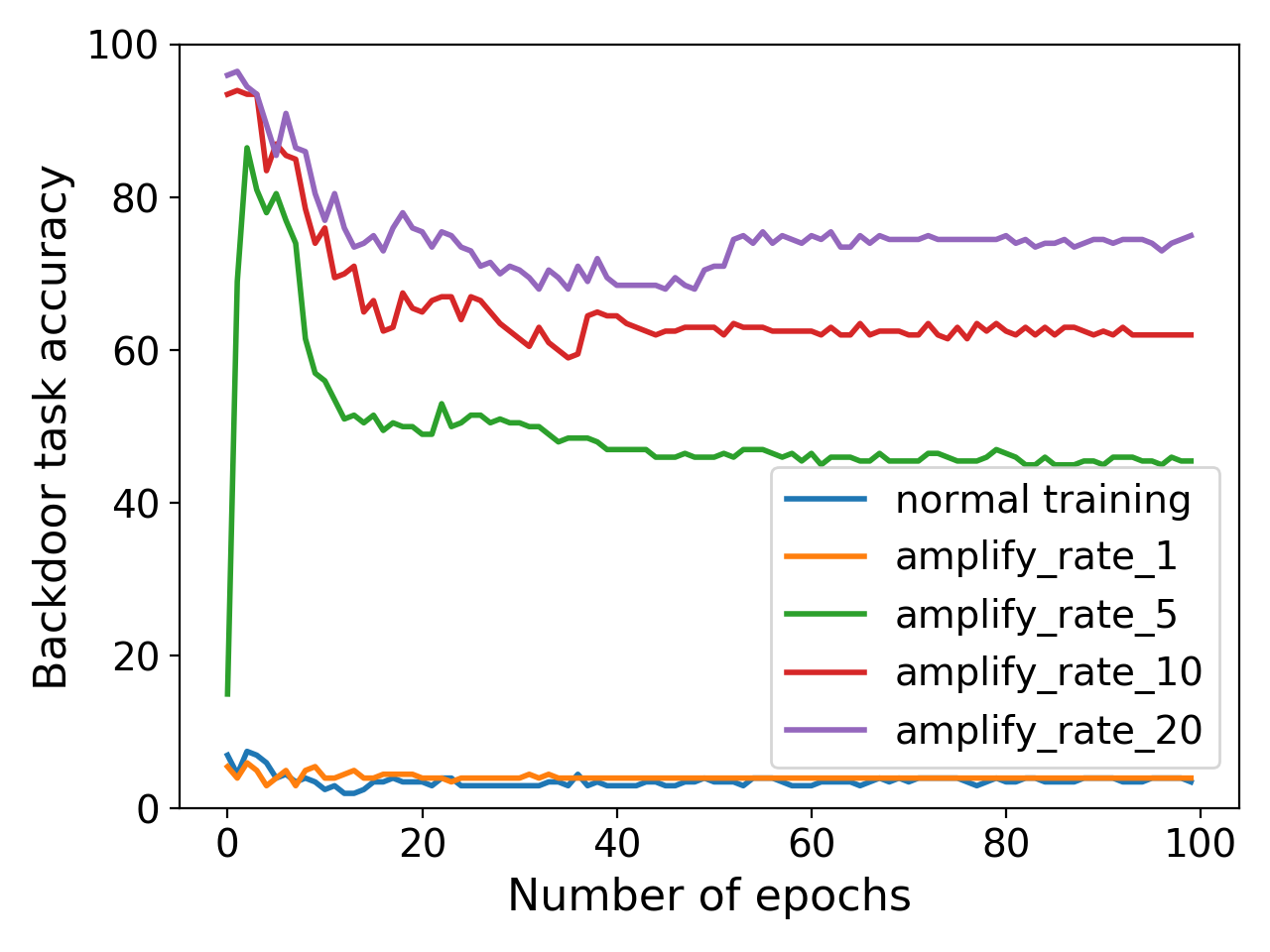}
\end{minipage}
}
\subfigure[CIFAR20]{
\begin{minipage}[t]{0.47\linewidth}
\centering
\includegraphics[width=1\linewidth]{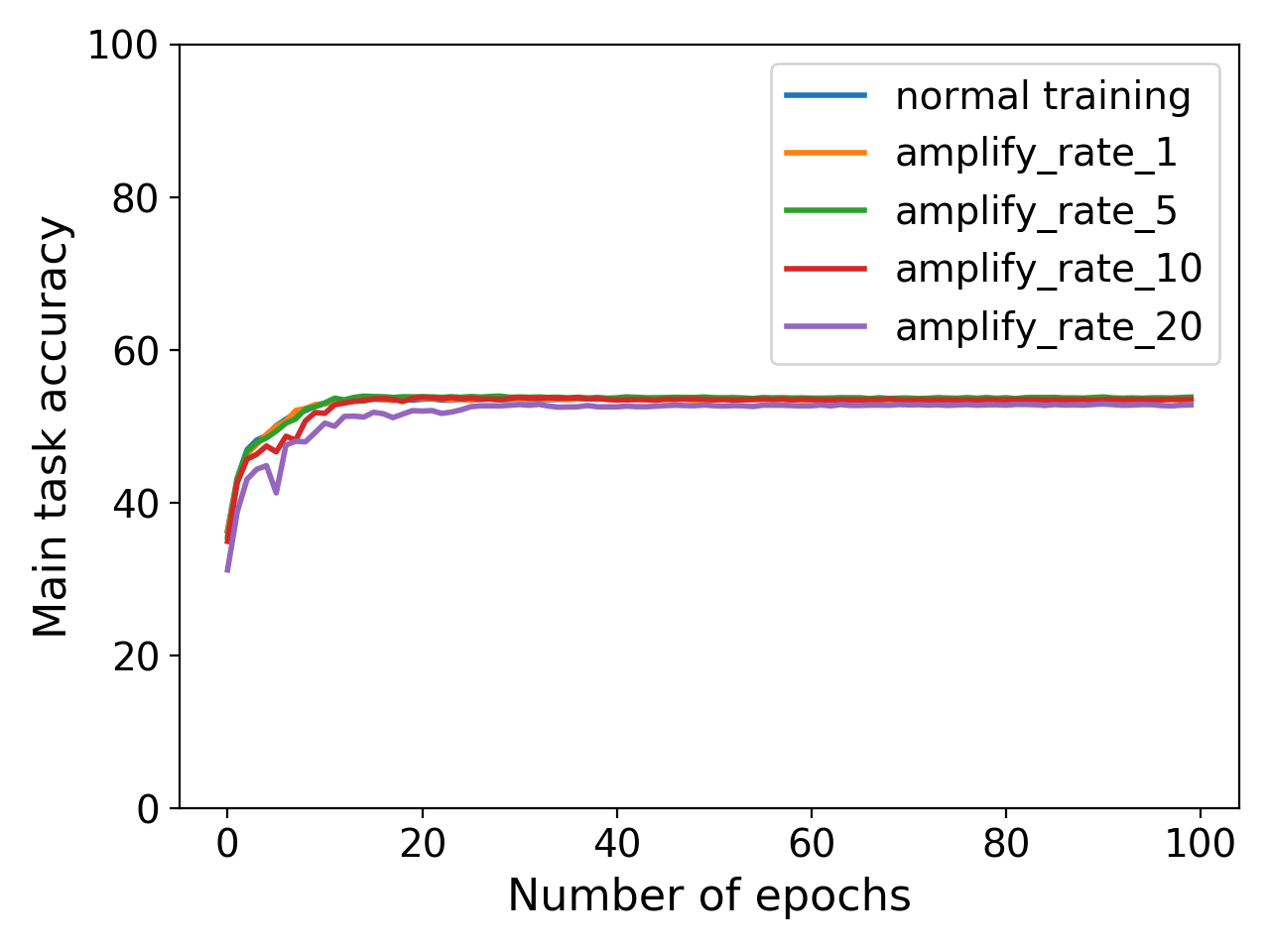}
\end{minipage}
}
\caption{Label replacement backdoor attacks on different datasets.}
\label{fig:acc_backdoor_training}
\end{figure}

For testing the effectiveness of the label inference attack, we use backdoor task success rate, i.e. backdoor accuracy, as our metric. If the backdoor sample is classified by the VFL system as the target class instead of its original class, the backdoor task is considered successful for that sample. The backdoor accuracy is then the ratio of successfully mislabeled backdoor samples. The accuracy of the main classification class is also taken into consideration. The results are shown in Figure\ref{fig:acc_backdoor_training}. The left column of Figure\ref{fig:acc_backdoor_training} shows the backdoor task accuracy at various levels of backdoor, indicated by the amplify rate $\gamma$. Some of which are more than $90\%$, meaning that the backdoor task has succeeded. The right column of Figure \ref{fig:acc_backdoor_training} shows the main task accuracy, which stays high for various attack levels. When $\gamma$ is set too large the backdoor task succeeded while the main task failed. In the following experiments, we set $\gamma$ to $10$, at which the main task and the backdoor task both work well.

\subsection{Defenses at Active Party}
\subsubsection{Defense Settings} 
We evaluate various defense mechanism on the previously proposed batch label inference attack: Differential Privacy (DP),  Gradient Sparsification (GS) and MARVELL \cite{label_leakage_VFL_2021}. Both encoder and decoder of the CoAE have the architecture:
\textbf{fc($(6C)^2$)}-\textbf{relu}-\textbf{fc($C$)}-\textbf{softmax}, where \textbf{fc} denotes fully-connected layer and $C$ denotes the class number. We evaluate the performance of CoAE with various $\lambda_2$ from $0.05$ to $2.0$.

\begin{figure}[!htb]
\centering
\subfigure[MNIST-LI]{
\begin{minipage}[t]{0.47\linewidth}
\centering
\includegraphics[width=1\linewidth]{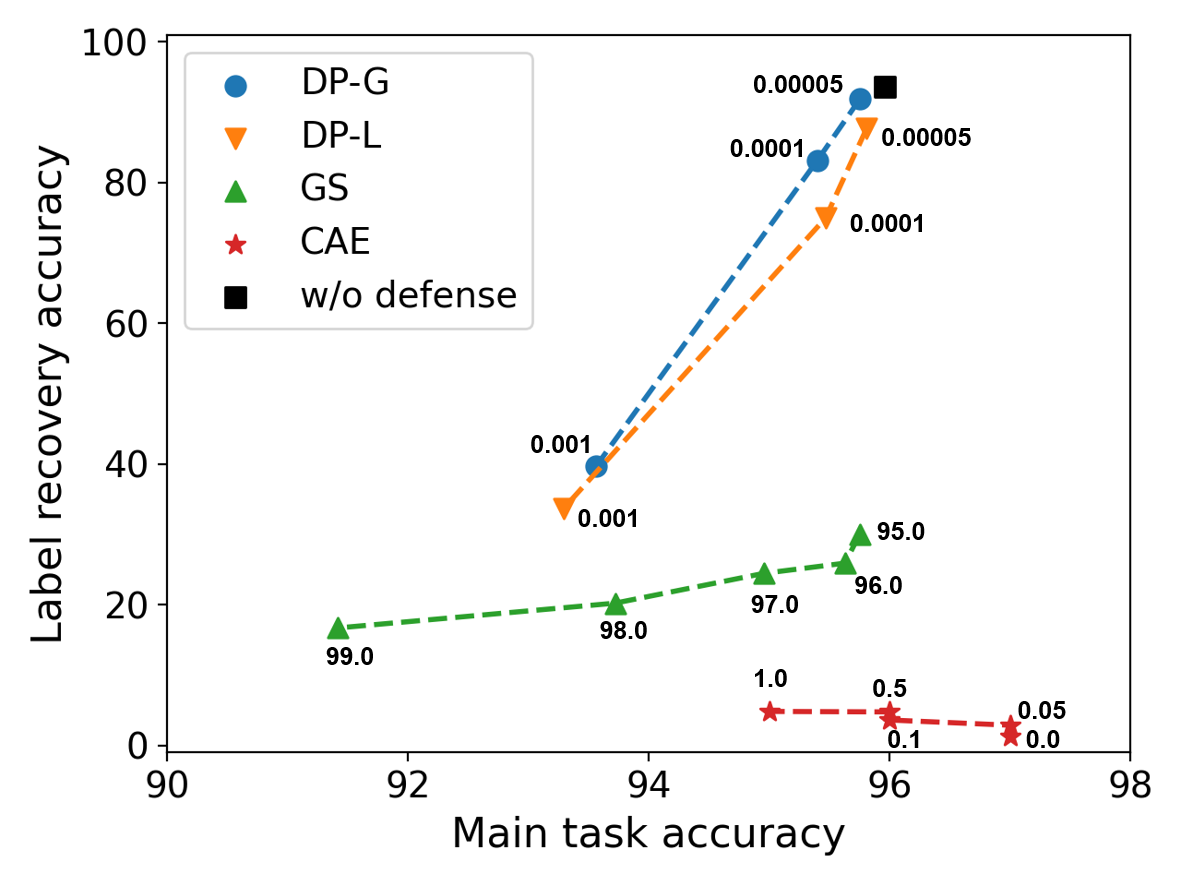}
\end{minipage}
}
\subfigure[MNIST-LI-binary]{
\begin{minipage}[t]{0.47\linewidth}
\centering
\includegraphics[width=1\linewidth]{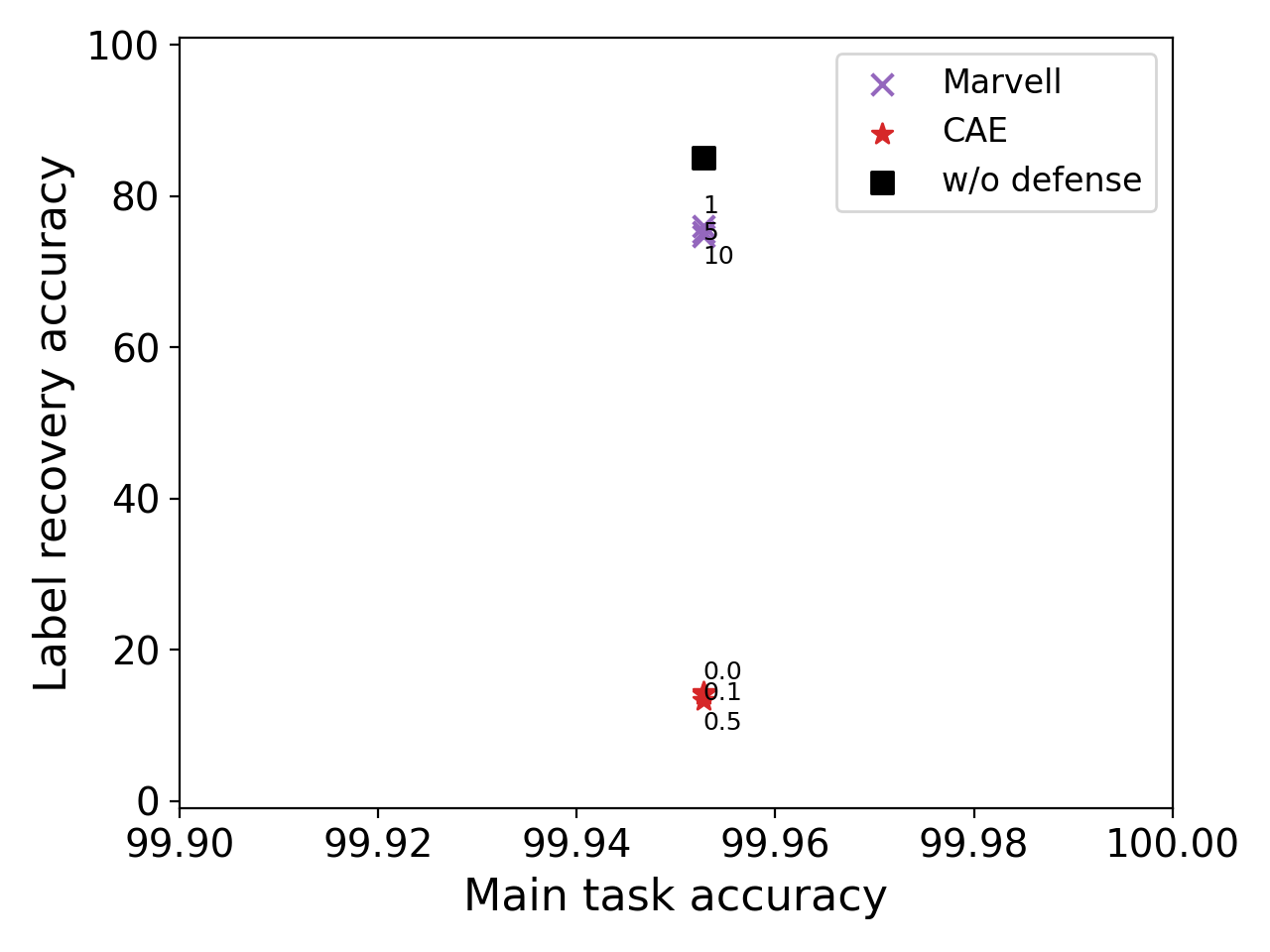}
\end{minipage}
}
\subfigure[NUSWIDE-LI]{
\begin{minipage}[t]{0.47\linewidth}
\centering
\includegraphics[width=1\linewidth]{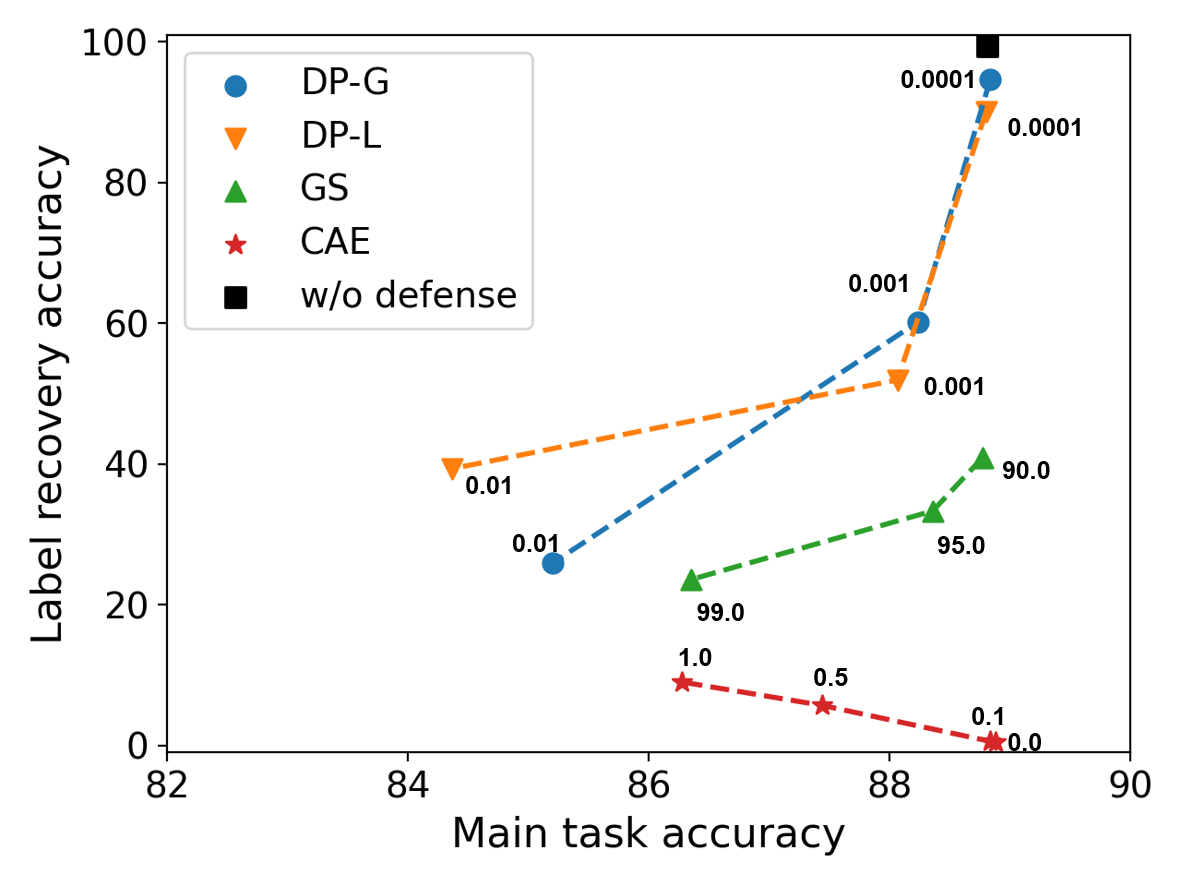}
\end{minipage}
}
\subfigure[NUSWIDE-LI-binary]{
\begin{minipage}[t]{0.47\linewidth}
\centering
\includegraphics[width=1\linewidth]{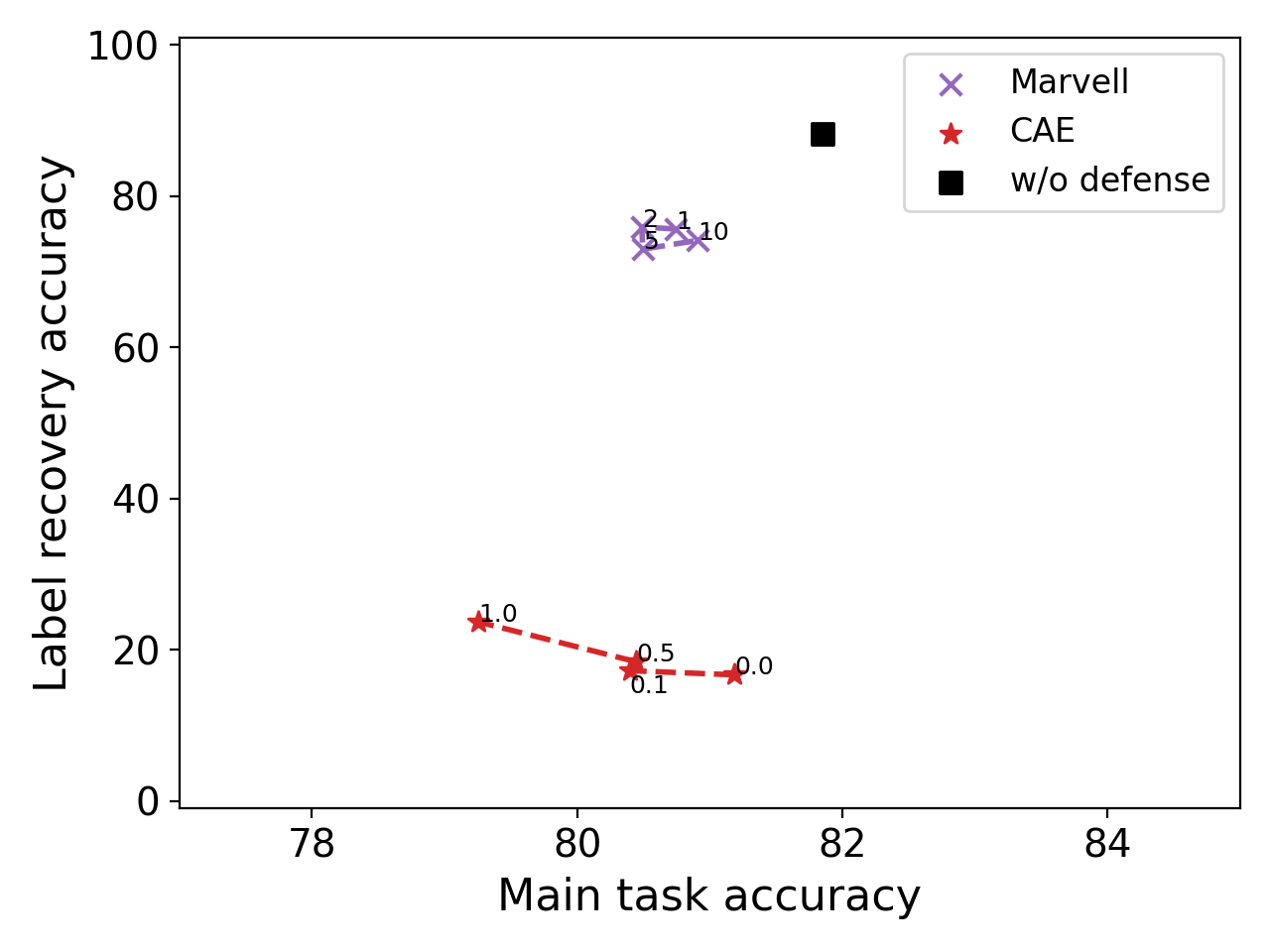}
\end{minipage}
}
\subfigure[CIFAR20-LI]{
\begin{minipage}[t]{0.47\linewidth}
\centering
\includegraphics[width=1\linewidth]{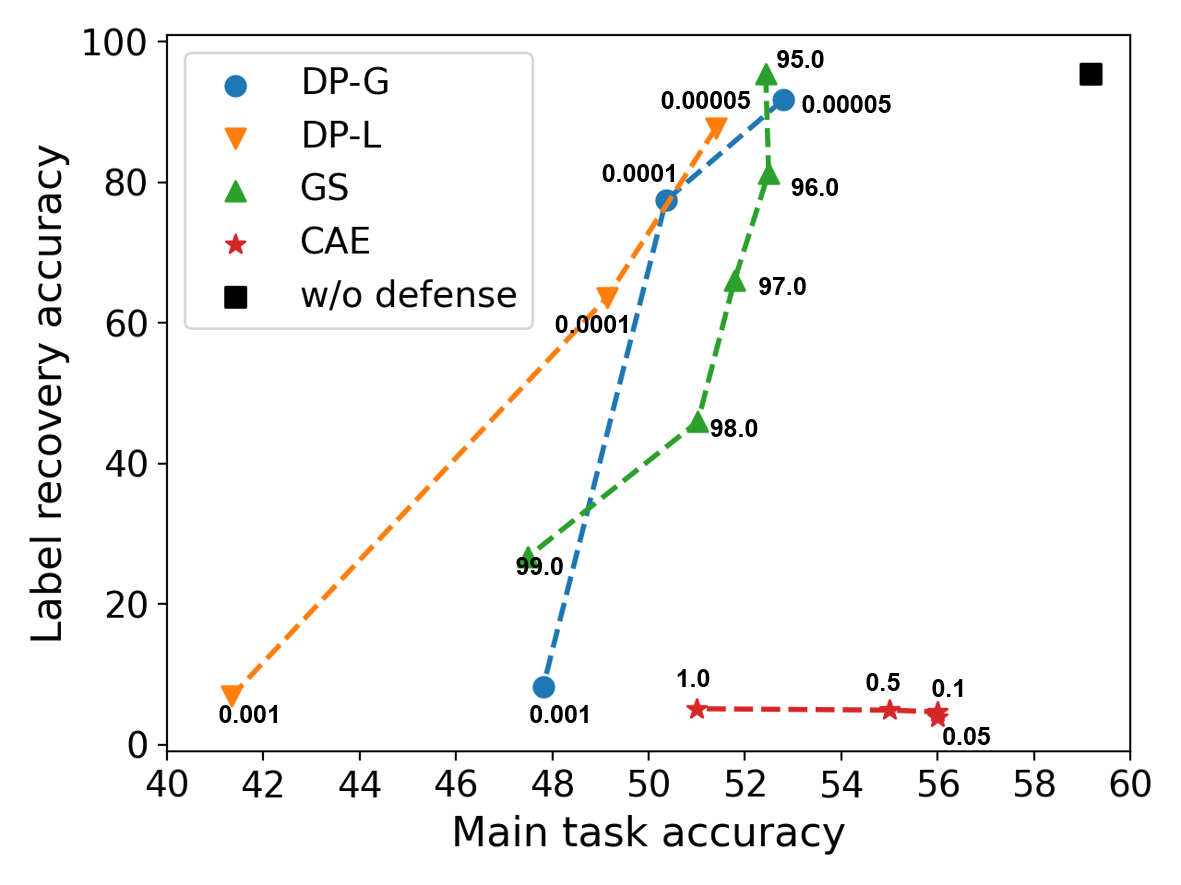}
\end{minipage}
}
\subfigure[CIFAR10-LI-binary]{
\begin{minipage}[t]{0.47\linewidth}
\centering
\includegraphics[width=1\linewidth]{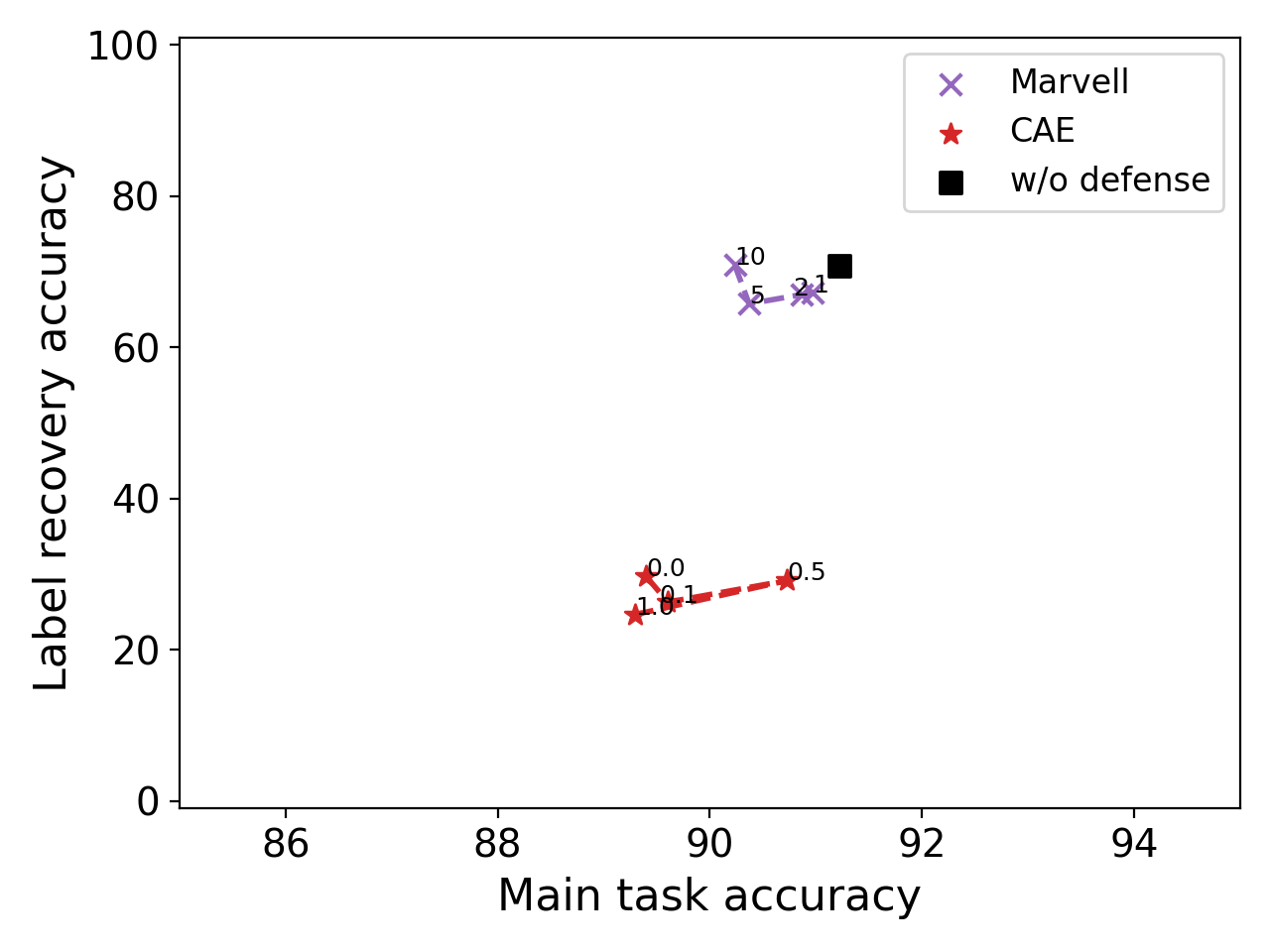}
\end{minipage}
}
\caption{Main task accuracy vs. batch label inference task (LI) accuracy under various defense strategies on multi-class classification task and binary classification task. The numbers on the figures are controlled variables (DP-G: $\sigma$, DP-L: $b$, GS: drop rate $s$, MARVELL: ratio of power constraint hyperparameter and gradient $s$, CoAE: $\lambda_2$)}
\label{fig:LI_defense_label_recovery}
\end{figure}

For DP, gaussian differential private mechanism (DP-G) and Laplacian (DP-L) differential private mechanism are employed. Gradients were $2$-norm clipped with $0.2$. Then, a Gaussian or Laplacian noise was added. The standard deviation of the applied noise ranges from $0.001$ to $0.1$. For GS, we evaluate various drop rates, from $99.0$ to $99.9$. For MARVELL, as it only supports binary classification, we evaluate its defense performance against our method, CoAE, separately using only the data from two classes. In our experiments, for MNIST and NUSWIDE datasets, we randomly selects two classes from the total dataset; while class  'clouds' and 'person' are used for CIFAR dataset in order to balance the number of data samples from those two classes. The power constraint hyperparameter various from $1$ to $10$ times the norm of gradients.  


\subsubsection{Defense Results} 
\textbf{Batch-level Label Inference Attack}\\The results of batch label inference defense of several strategies are shown in Figure \ref{fig:LI_defense_label_recovery}. The left column of Figure  \ref{fig:LI_defense_label_recovery} shows the results of defense under multi-class classification task (10, 5, 20 classes for MNIST, NUSWIDE and CIFAR separately), while the right column shows the results of binary classification task. In each figure, the upper left indicates high attack performance and low main task accuracy, so a better defense should be towards the lower right. Compared to other approaches, we see that our CoAE defense consistently achieves better trade-off performance, with high main task accuracy and low attack task accuracy for multi-class classification task, and relative the same main task accuracy and low attack accuracy for binary classification task. The above proves that CoAE is superior in defending batch label inference attack. Also, for this task, GS has a better performance than DP-based methods. And a change in the power constraint hyperparameter for MARVELL do not lead to a significant variation in the result.

\begin{center}
\begin{table*}[htb]
\centering
\begin{tabular}{m{2.5cm}<{\centering}|m{0.6cm}<{\centering}m{0.6cm}<{\centering}m{0.6cm}<{\centering}m{0.6cm}<{\centering}m{0.6cm}<{\centering}m{0.6cm}<{\centering}m{0.6cm}<{\centering}m{0.6cm}<{\centering}m{0.6cm}<{\centering}m{0.6cm}<{\centering}m{0.6cm}<{\centering}m{0.6cm}<{\centering}m{0.6cm}<{\centering}m{0.6cm}<{\centering}}
\toprule[1.2pt]
\multirow{2}*{Attack} & \multicolumn{2}{c}{\shortstack{No\\Defense}} & \multicolumn{2}{c}{\shortstack{NG (DP-L)}} & \multicolumn{2}{c}{\shortstack{GC (GS)}} & \multicolumn{2}{c}{\shortstack{PPDL}} & \multicolumn{2}{c}{\shortstack{D-SGD}} & \multicolumn{2}{c}{\shortstack{CoAE}} & \multicolumn{2}{c}{\shortstack{CoAE +\\D-SGD}} \\
\cline{2-15}
\specialrule{0em}{0pt}{1pt}
~ & Attack & Main Task & Attack & Main Task & Attack & Main Task & Attack & Main Task & Attack & Main Task & Attack & Main Task & Attack & Main Task  \\ 
\midrule[1.3pt]
\multirow{3}*{\shortstack{Passive Model\\Completion\cite{fulabel}\\(40 auxiliary data)}} & \multirow{3}*{0.698} & \multirow{3}*{0.811} & \multirow{3}*{0.306} & \multirow{3}*{0.517} & \multirow{3}*{0.370} & \multirow{3}*{0.543} & \multirow{3}*{0.447} & \multirow{3}*{0.687} & \multirow{3}*{0.252} & \multirow{3}*{0.671} & 0.642 & 0.797 & 0.280 & 0.737\\
~ & ~ & ~ & ~ & ~ & ~ & ~ & ~ & ~ & ~ & ~ & 0.661 & 0.801 & 0.317 & 0.793\\
~ & ~ & ~ & ~ & ~ & ~ & ~ & ~ & ~ & ~ & ~ & 0.648 & 0.804 & 0.537 & 0.801\\
\specialrule{0em}{2pt}{0pt}
\specialrule{0em}{0pt}{3pt}
\multirow{3}*{\shortstack{Passive Model\\Completion\cite{fulabel}\\(10 auxiliary data)}} & \multirow{3}*{0.618} & \multirow{3}*{0.811} & \multirow{3}*{0.282} & \multirow{3}*{0.517} & \multirow{3}*{0.213} & \multirow{3}*{0.543} & \multirow{3}*{0.426} & \multirow{3}*{0.687} & \multirow{3}*{0.214} & \multirow{3}*{0.671} & 0.605 & 0.797 & 0.254 & 0.737\\
~ & ~ & ~ & ~ & ~ & ~ & ~ & ~ & ~ & ~ & ~ & 0.616 & 0.801 & 0.278 & 0.793\\
~ & ~ & ~ & ~ & ~ & ~ & ~ & ~ & ~ & ~ & ~ & 0.606 & 0.804 & 0.462 & 0.801\\
\specialrule{0em}{2pt}{0pt}
\specialrule{0em}{0pt}{3pt}
\multirow{3}*{\shortstack{Active Model\\Completion\cite{fulabel}\\(40 auxiliary data)}} & \multirow{3}*{0.741} & \multirow{3}*{0.808} & \multirow{3}*{0.531} & \multirow{3}*{0.618} & \multirow{3}*{0.537} & \multirow{3}*{0.574} & \multirow{3}*{0.397} & \multirow{3}*{0.666} & \multirow{3}*{0.231} & \multirow{3}*{0.660} & 0.721 & 0.800 & 0.550 & 0.674\\ 
~ & ~ & ~ & ~ & ~ & ~ & ~ & ~ & ~ & ~ & ~ & 0.722 & 0.801 & 0.640 & 0.775\\
~ & ~ & ~ & ~ & ~ & ~ & ~ & ~ & ~ & ~ & ~ & 0.717 & 0.810 & 0.687 & 0.802\\
\specialrule{0em}{2pt}{0pt}
\specialrule{0em}{0pt}{3pt}
\multirow{3}*{\shortstack{Active Model\\Completion\cite{fulabel}\\(10 auxiliary data)}} & \multirow{3}*{0.658} & \multirow{3}*{0.808} & \multirow{3}*{0.413} & \multirow{3}*{0.618} & \multirow{3}*{0.235} & \multirow{3}*{0.574} & \multirow{3}*{0.215} & \multirow{3}*{0.666} & \multirow{3}*{0.228} & \multirow{3}*{0.660} & 0.625 & 0.800 & 0.359 & 0.674\\ 
~ & ~ & ~ & ~ & ~ & ~ & ~ & ~ & ~ & ~ & ~ & 0.643 & 0.801 & 0.391 & 0.775\\
~ & ~ & ~ & ~ & ~ & ~ & ~ & ~ & ~ & ~ & ~ & 0.649 & 0.810 & 0.648 & 0.802\\
\specialrule{0em}{2pt}{0pt}
\specialrule{0em}{0pt}{3pt}
\multirow{3}*{\shortstack{Direct Label\\Inference\cite{fulabel}}} & \multirow{3}*{1.000} & \multirow{3}*{0.831} &  \multirow{3}*{0.491} & \multirow{3}*{0.832} & \multirow{3}*{0.006} & \multirow{3}*{0.100} & \multirow{3}*{0.389} & \multirow{3}*{0.826} & \multirow{3}*{0.940} & \multirow{3}*{0.747} & 0.000 & 0.805 & 0.000 & 0.795\\ 
~ & ~ & ~ & ~ & ~ & ~ & ~ & ~ & ~ & ~ & ~ & 0.001 & 0.818 & 0.000 & 0.810\\
~ & ~ & ~ & ~ & ~ & ~ & ~ & ~ & ~ & ~ & ~ & 0.001 & 0.828 & 0.000 & 0.826\\
\specialrule{0em}{2pt}{0pt}
\hline
\specialrule{0em}{0pt}{3pt}
\multirow{3}*{\shortstack{Gradient\\Inversion\\(this work)}} & \multirow{3}*{0.893} & \multirow{3}*{0.815} & \multirow{3}*{0.144} & \multirow{3}*{0.469} & \multirow{3}*{0.874} & \multirow{3}*{0.182} & \multirow{3}*{0.177} & \multirow{3}*{0.592} & \multirow{3}*{0.890} & \multirow{3}*{0.711} & 0.247 & 0.803 & 0.060 & 0.725\\ 
~ & ~ & ~ & ~ & ~ & ~ & ~ & ~ & ~ & ~ & ~ & 0.263 & 0.806 & 0.060 & 0.743\\
~ & ~ & ~ & ~ & ~ & ~ & ~ & ~ & ~ & ~ & ~ & 0.297 & 0.804 & 0.061 & 0.791\\
\bottomrule[1.2pt]
\end{tabular}
\caption{Comparison of defense and attack results of different label inference attacks in VFL setting. Experiments are carried out on CIFAR10 dataset, using all the 10 classes of data. For our gradient inversion attack, batch size is $2048$. The value reported in the table are attack accuracy and model accuracy respectively. We chose noise level at $0.001$ for Noisy Gradients (the same attack as DP-L), compression rate at $0.1$ for Gradient Compression (the same attack as GS), fraction of gathered gradient values at $0.5$ for Privacy-preserving Deep Learning, $12$ bins for Discrete SGD and $\lambda=1.0, 0.5, 0.0$ respectively for the three rows of results for CoAE.}
\label{tab:label_inference_comparison_2}
\end{table*}
\end{center}

In Table \ref{tab:label_inference_comparison_2}, we also compare the effectiveness of the four defense methods in \cite{fulabel}, including Noisy Gradients (NG), Gradient Compression (GC), Privacy-Preserving Deep Learning (PPDL), Discrete SGD (D-SGD) and our defensing method CoAE, in defending various label inference attacks. NG and GC are the same defense method sas laplace-DP (DP-L) and GC respectively, two of the defense methods we test above. We choose CIFAR10 as the dataset for experiments. For our gradient inversion attack, we adopt a batch size of 2048, whereas for model completion attacks, 40 or 10 auxiliary labels are used. It's clear that our method CoAE outperforms all the other defensing methods for both batch-level and sample-level label inference attack with a low attack accuracy and high main task accuracy, whereas Discrete SGD method performs better at defending model completion tasks but is not effective for gradient inversion tasks. This is reasonable since Discrete SGD keeps the sign of gradients to maintain its main task accuracy which can still be exploited for label inference attacks with both sample-level and batch-level gradients. On the other hand, CoAE is not designed for scenarios where sufficient prior knowledge about labels can be exploited therefore are not effective for defending model completion tasks. Since the attacker in model completion attacks knows additional auxiliary data for each class that include both the data and the real labels, when the attacker fine-tunes its model, it can learn the relationship between its original predictions learned in the VFL training period and the real labels. To take full advantages of both methods, we propose to combine coAE with Discrete SGD methods for a universal strategy to defend label inference attacks in VFL. As discussed earlier, our proposed coAE is orthogonal to other information-reduction methods therefore can be easily combined with them. Specifically, when combining with Discrete SGD, after using CoAE to replace original real labels with fake soft labels, the training and defending process is just the same as they are when only Discrete SGD is applied. As shown in the Table \ref{tab:label_inference_comparison_2}, the proposed combination method exhibits a better trade-off between main task and attack accuracy for both model completion and gradient inversion attacks compared to standalone Discrete SGD and COAE respectively. 

To sum up, our CoAE defense is the most effective method for blocking both batch level and sample level label inference attack for VFL scenarios, when other defense methods perform unsatisfactorily, and can be readily combined with discrete SGD method to block passive and active model completion attacks, where it helps to increase the main task accuracy to a large extent with a slightly negative impact on the attack accuracy.\\

\begin{figure}[!htb]
\centering
\subfigure[MNIST-LR]{
\begin{minipage}[t]{0.47\linewidth}
\centering
\includegraphics[width=1\linewidth]{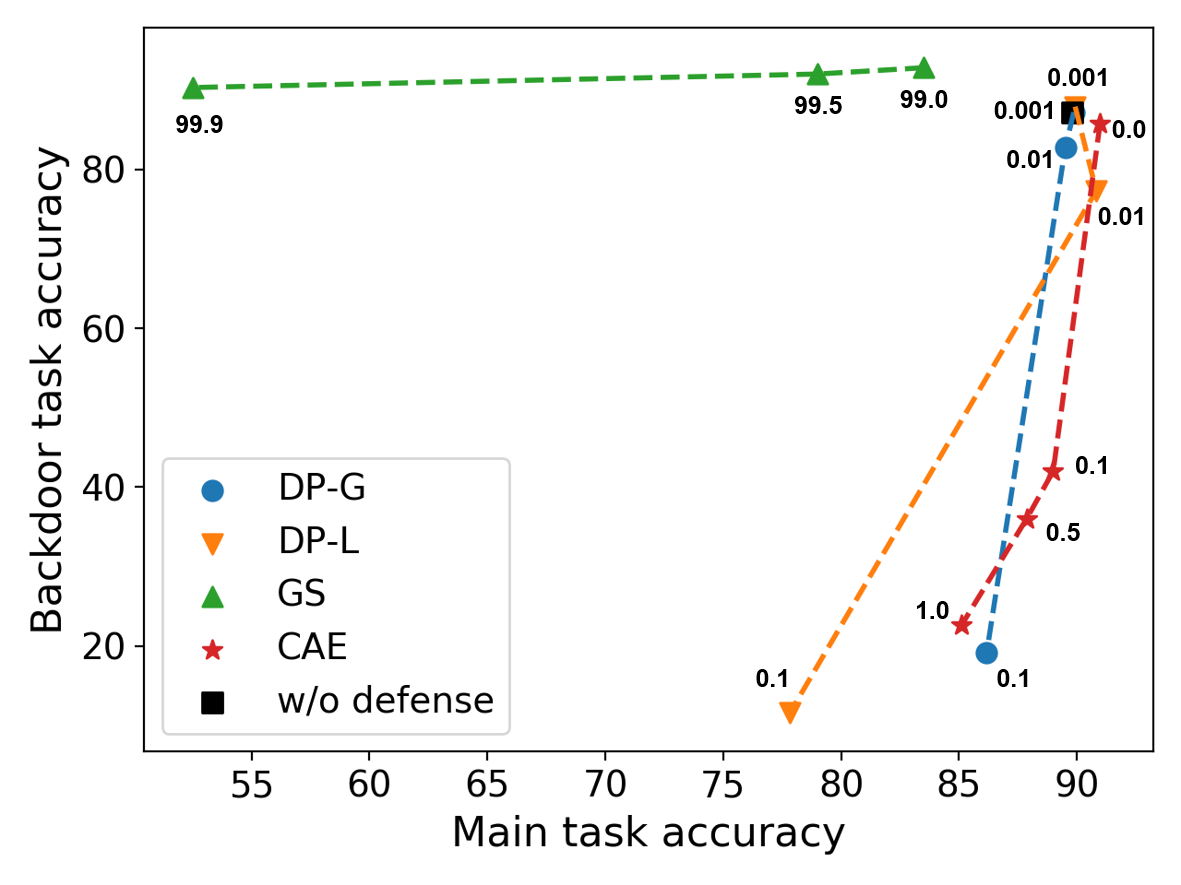}
\end{minipage}
}
\subfigure[NUSWIDE-LR]{
\begin{minipage}[t]{0.47\linewidth}
\centering
\includegraphics[width=1\linewidth]{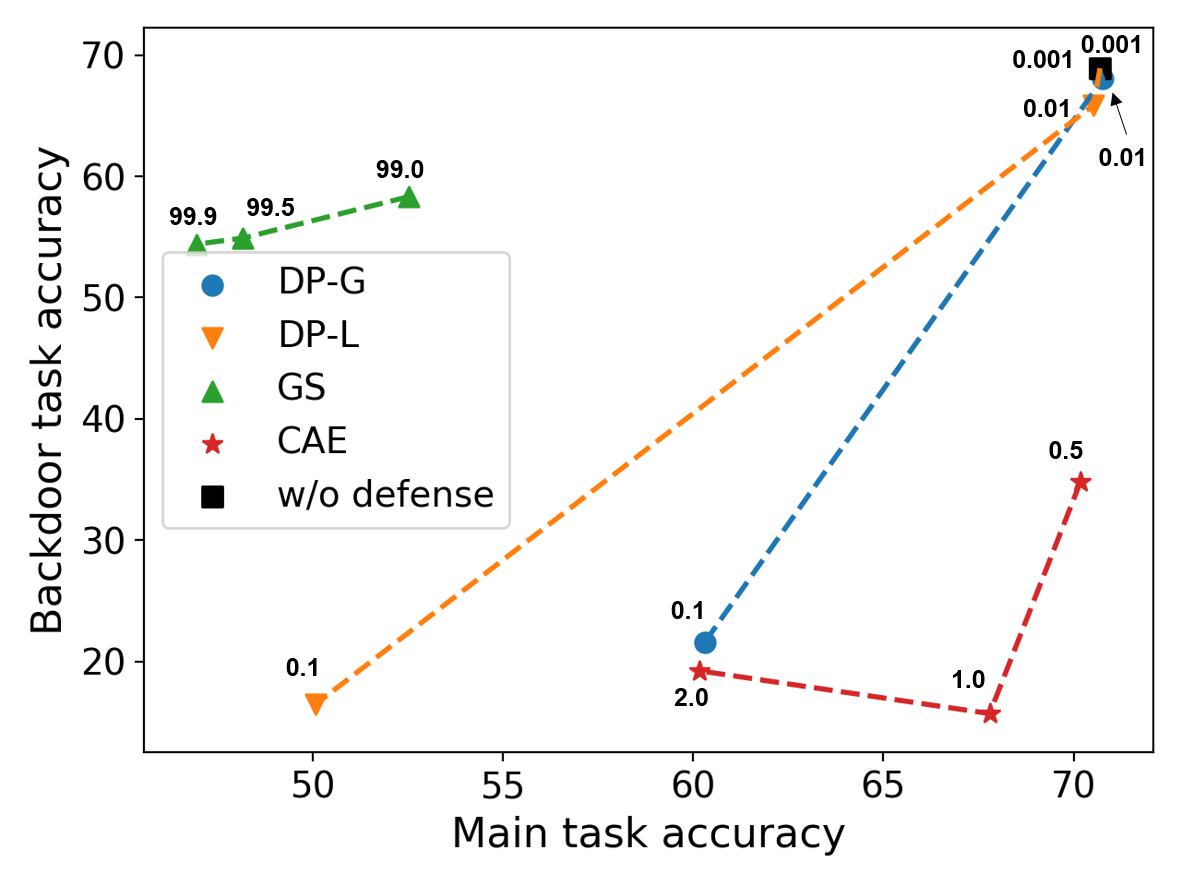}
\end{minipage}
}
\subfigure[CIFAR20-LR]{
\begin{minipage}[t]{0.47\linewidth}
\centering
\includegraphics[width=1\linewidth]{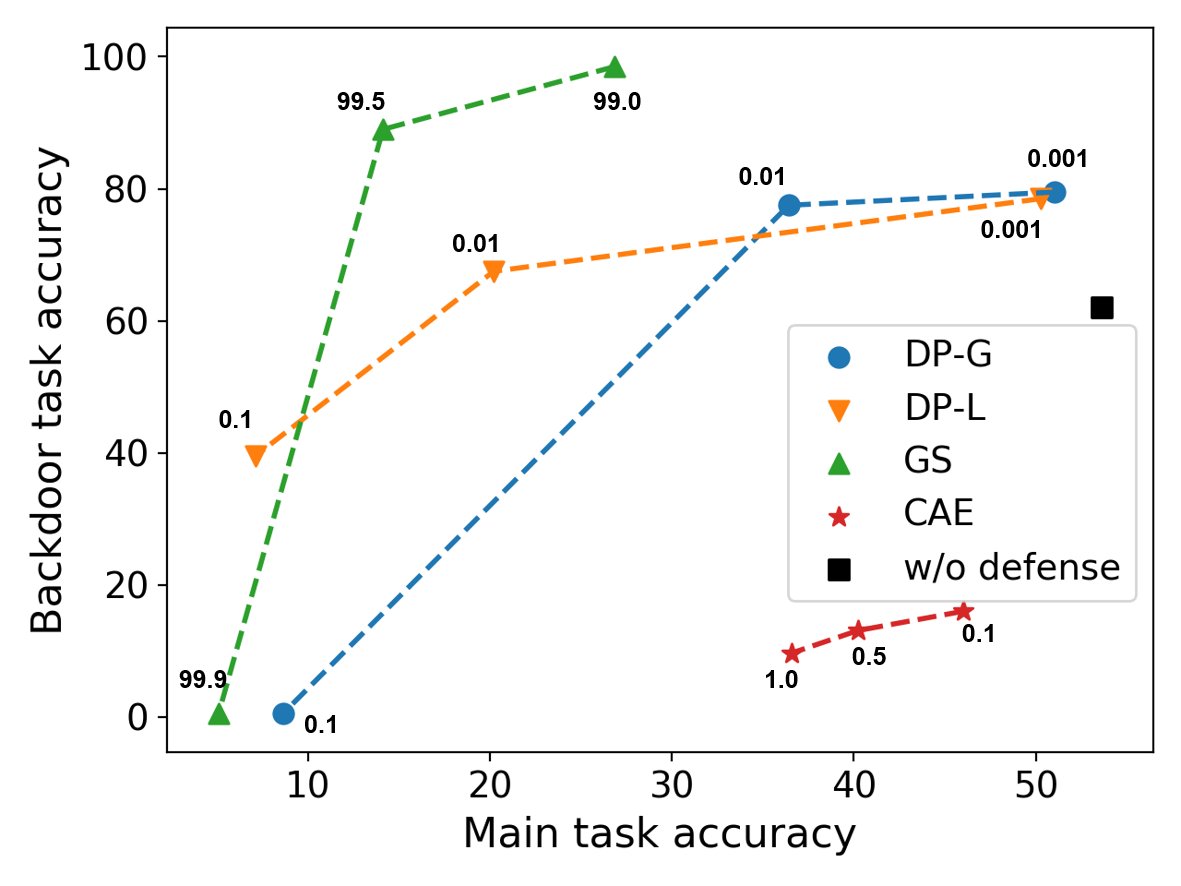}
\end{minipage}
\label{fig:GR_singleparty_defense_backdoor}
}
\subfigure[CIFAR20-LR-Distributed]{
\begin{minipage}[t]{0.47\linewidth}
\centering
\includegraphics[width=1\linewidth]{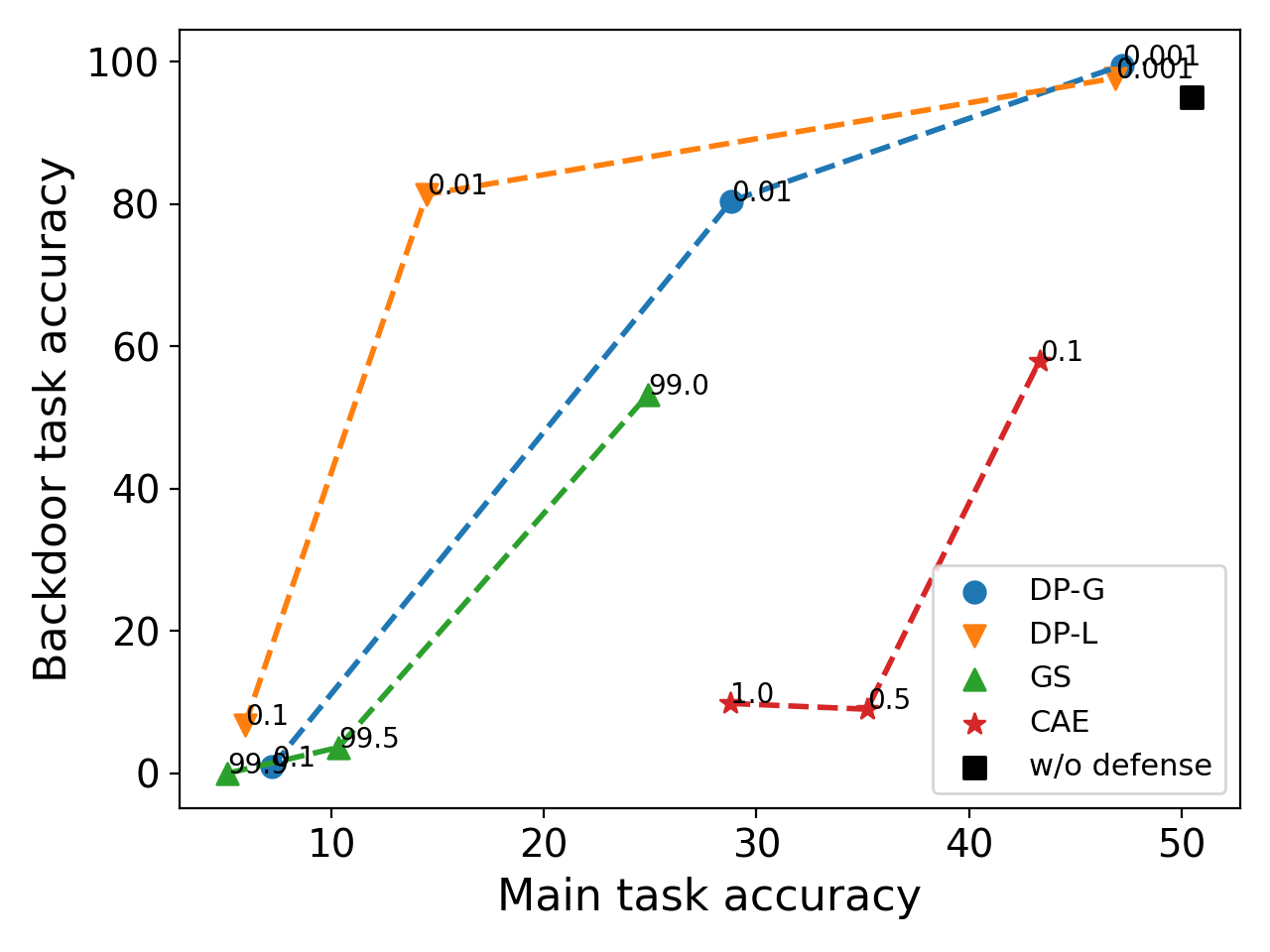}
\end{minipage}
\label{fig:GR_Distributed_defense_backdoor}
}
\caption{Main task accuracy vs. Label-Replacement backdoor (GR) accuracy under various defense strategies. The numbers on the figures are controlled variables (DP-G: $\sigma$, DP-L: $b$, GS: drop rate $s$, CoAE: $\lambda_2$)}
\label{fig:GR_defense_backdoor}
\end{figure}

\noindent\textbf{Label Replacement Backdoor Attack}\\We also evaluate our proposed label defense mechanism on gradient-replacement attack and compared our methods with two baselines: Differential Privacy (DP) and Gradient Sparsification (GS). Both encoder and decoder of the CoAE have the same architecture as displayed above. The hyper-parameters for the two baseline methods are the same as for label inference attack as well. The results are shown in Figure \ref{fig:GR_defense_backdoor}.

This experiment is carried out under multi-class classification task setting (10, 5, 20 classes for MNIST, NUSWIDE and CIFAR separately). Same as the batch label inference attack, a better defense should be towards the lower right in each figure. It's clear to see from Figure \ref{fig:GR_defense_backdoor} that results of gradient sparsification (GS) are in the upper left corner, indicating a high backdoor accuracy and low main task accuracy. This suggests that GS is inferior to defend our backdoor attack. Comparing to DP and GS mechanism, our CoAE method can defend backdoor attack to the same level, while retaining a high accuracy of the main task. 

We further evaluate our defense strategy under distributed backdoor attack settings as the one proposed in previous work \cite{xie2020dba}. We compare our method, CoAE, with three baseline methods on CIFAR100 dataset astill using only images from 20 classes. In this experiment, the feature of data is equally partitioned into four parties with only one active party owning the labels and three other passive parties without label information. Similar as previous work \cite{xie2020dba}, the three passive parties work together and conduct gradient-replacement-backdoor attack to the active party. Each attacker has their one-pixel trigger at the lower right corner of each data sample of their own which together forming a three-pixel trigger. Notice that trigger is smaller than the four-pixel trigger we use in previous experiments mentioned above. The result is shown in Figure \ref{fig:GR_Distributed_defense_backdoor}. It's clear that our method beats all the baseline methods, achieving a low backdoor task accuracy at a high main task accuracy. Also, comparing with Figure \ref{fig:GR_singleparty_defense_backdoor}, although distributed backdoor attack is much stronger than single-party backdoor attack (comparing the two black square in each plot which demonstrate the pure attacking results without defense), our defense strategy as well as the baseline strategies can achieve similar defending results.

From Figure \ref{fig:GR_singleparty_defense_backdoor} and Figure \ref{fig:GR_Distributed_defense_backdoor}, we can see that in label-replacement backdoor attack, as the noise level of DP or the drop rate of GS increases, backdoor accuracy decreases at the expense of significant drop in main task accuracy. Notice that the trends in the performance of CoAE in label-replacement backdoor is the opposite to that in label inference attacks. For backdoor attacks, backdoor accuracy decreases as the confusion level grows, indicating confusion is important for defending the attack. Without confusion ($\lambda_2=0$), the backdoor attack will still succeed since then the autoencoder's sole function is to switch labels among classes and the transformation would work the same for the backdoor samples and other samples.

\begin{figure}[!htb]
\centering
\subfigure[CoAE without entropy loss]{
\label{wo_entr_loss}
\begin{minipage}[t]{0.472\linewidth}
\centering
\includegraphics[width=1\linewidth]{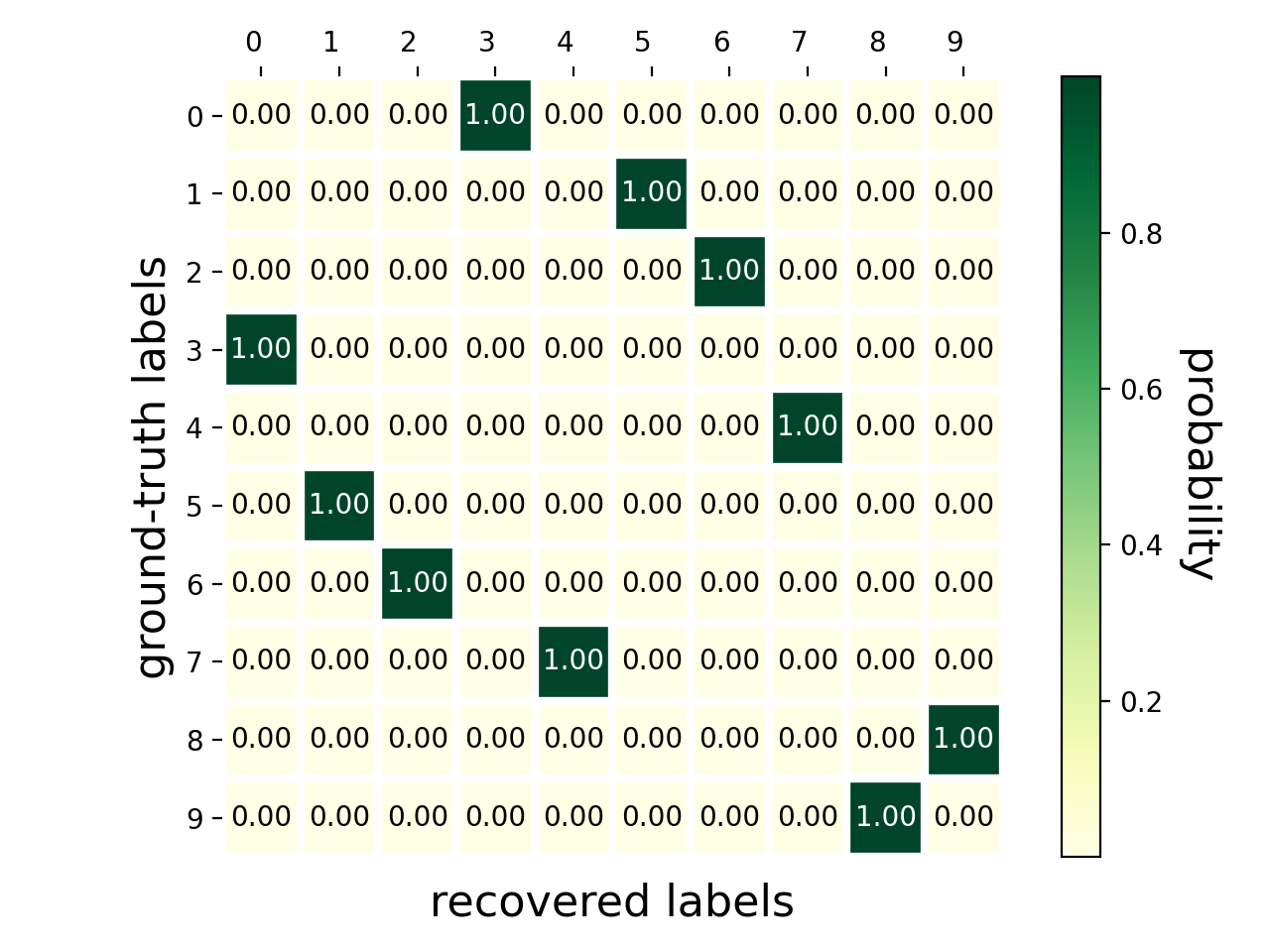}
\end{minipage}}
\subfigure[CoAE with entropy loss]{
\label{w_entr_loss}
\begin{minipage}[t]{0.472\linewidth}
\centering
\includegraphics[width=1\linewidth]{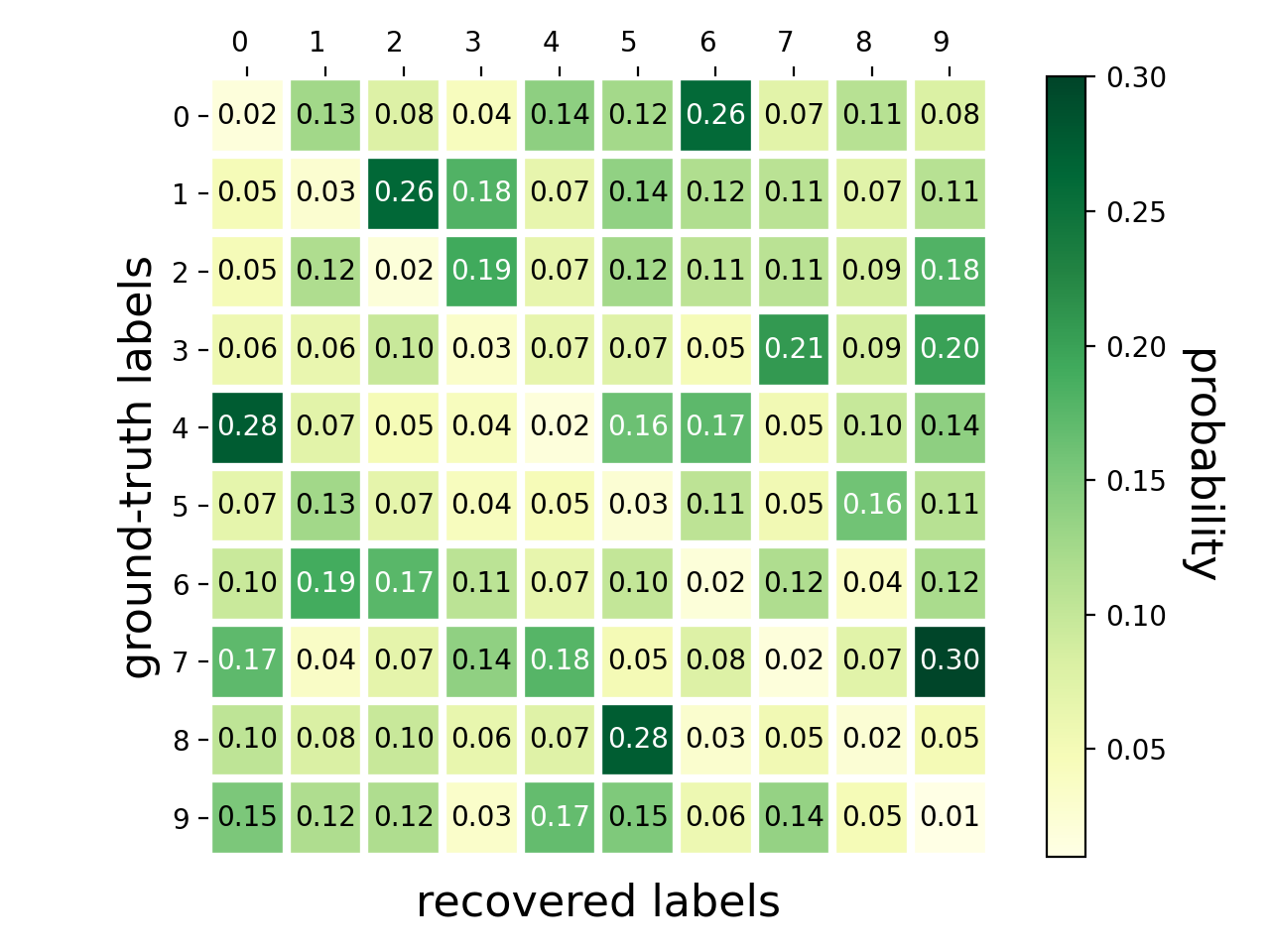}
\end{minipage}}
\caption{The probability distribution (PD) over labels restored by the passive party when the CoAE defense is applied at the active party. (a) PD matrix when CoAE is trained without entropy loss (i.e. $\lambda_2=0$). (b) PD matrix when CoAE is trained with entropy loss (i.e. $\lambda_2=1$).}
\label{fig:ae_entropy}
\end{figure}

\subsubsection{Impact of Confusion}

In this section, we analyze the impact of confusion coefficient. In defending batch label inference attacks, as the confusion coefficient increases, the label recovery rate gradually goes up, because higher confusion means higher probability for the soft fake labels to be mapped into the true labels.  




Figure \ref{fig:ae_entropy} depicts the probability distribution (PD) over labels restored by the passive party conducting batch label inference attack (Algo. \ref{label_leakage_algorithm}) on CIFAR10, when the active party is performing CoAE protection (Algo. \ref{algo:coAE}). Specifically, each element in the PD matrix is computed by $\frac{C_{ij}}{\sum_j C_{ij}}$, where $C_{ij}$ denotes the number of samples having ground-truth label $i$ but is restored with label $j$. Figure \ref{w_entr_loss} and Figure 
\ref{wo_entr_loss} depict the probability distribution when the CoAE is trained with ($\lambda_2=1$) and without ($\lambda_2=0$) entropy loss $L_{entropy}$, respectively. Without entropy loss, the PD matrix is sparse so the attacker can learn all samples having the same label belong to the same class. With confusion, samples belonging to the same class are restored as multiple alternative labels, demonstrating that the passive party cannot classify its samples based on restored labels.



\section{Conclusions and Future Work}
We systematically study the batch-level label inference and replacement attacks for black-boxed feature-partitioned federated learning (VFL) problem for the first time, and propose gradient-inversion and gradient-replacement attacks at passive party without the help of either auxiliary labeled data or per-sample gradients. We analytically show that the labels can be fully recovered when batch size is small. Then, we propose confusional autoencoder (CoAE) as a novel defense strategy against these attacks. Our experiments testify that when implementing our CoAE defense mechanism, batch label inference attack can be prevented with superior performance over existing methods without noticeable changes to the other parties and the training protocol. However, CoAE is not designed for attacks with sufficient auxiliary information, in which case we show the combination of CoAE and information-reduction techniques, such as discrete SGD, can achieve a better trade-off between main accuracy and attack accuracy. Since CoAE defends well attacks exploiting a direct connection through gradient propagation, its usefulness when combining with other defense techniques deserves future exploration.       




\bibliographystyle{plain}
\bibliography{ref}

\end{document}